\documentclass[10pt,twocolumn,letterpaper]{article}

\usepackage[pagenumbers]{cvpr} 

\usepackage{graphicx}
\usepackage{amsmath}
\usepackage{amssymb}

\usepackage[subrefformat=parens]{subcaption}
\usepackage{float}
\usepackage{afterpage}
\usepackage{comment}
\usepackage{caption}
\usepackage{booktabs}       
\usepackage{longtable}
\usepackage{array}
\usepackage{multirow}
\usepackage[accsupp]{axessibility}  

\usepackage[pagebackref,breaklinks,colorlinks]{hyperref}

\usepackage[capitalize]{cleveref}
\crefname{section}{Sec.}{Secs.}
\Crefname{section}{Section}{Sections}
\Crefname{table}{Table}{Tables}
\crefname{table}{Tab.}{Tabs.}

\begin{document}

\title{Neural RGB-D Surface Reconstruction}

\author{
Dejan Azinovi{\'c}$^1$~~~
Ricardo Martin-Brualla$^2$~~
Dan B Goldman$^2$~~
Matthias Nie{\ss}ner$^1$~~~
Justus Thies$^{1,3}$~~~
\vspace{0.4cm} \\ 
$^1$Technical University of Munich~~~
$^2$Google Research~~~
$^3$Max Planck Institute for Intelligent Systems~~~
}


\twocolumn[{%
\renewcommand\twocolumn[1][]{#1}%
\maketitle
\begin{center}
	\includegraphics[width=\linewidth]{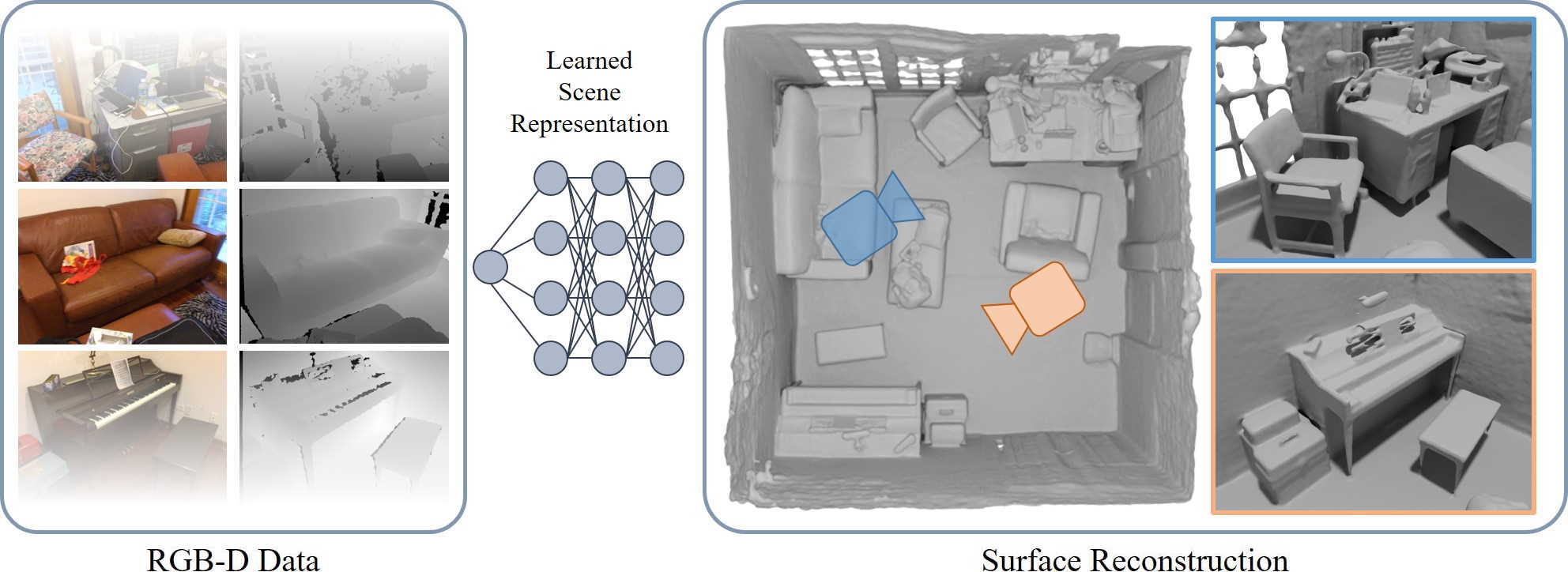}
	\vspace{-0.5cm}
    \captionof{figure}{
    Our method obtains a high-quality 3D reconstruction from an RGB-D input sequence by training a multi-layer perceptron. The core idea is to reformulate the neural radiance field definition in NeRF~\cite{mildenhall2020nerf}, and replace it with a differentiable rendering formulation based on signed distance fields which is specifically tailored to geometry reconstruction. 
    }
\label{fig:teaser}
\end{center}
}]


\newcommand{\x}{\mathbf{x}} 	
\newcommand{\y}{\mathbf{y}} 	
\newcommand{\n}{\mathbf{n}} 	
\newcommand{\wo}{\mathbf{i}}	
\newcommand{\wi}{\mathbf{o}}	
\newcommand{\X}{\mathbf{X}} 	

\newcolumntype{L}[1]{>{\raggedright\let\newline\\\arraybackslash\hspace{0pt}}m{#1}}
\newcolumntype{C}[1]{>{\centering\let\newline\\\arraybackslash\hspace{0pt}}m{#1}}
\newcolumntype{R}[1]{>{\raggedleft\let\newline\\\arraybackslash\hspace{0pt}}m{#1}}

\renewcommand{\paragraph}[1]{\medskip\noindent\textbf{#1}}
\begin{abstract}
Obtaining high-quality 3D reconstructions of room-scale scenes is of paramount importance for upcoming applications in AR or VR.
These range from mixed reality applications for teleconferencing, virtual measuring, virtual room planing, to robotic applications.
While current volume-based view synthesis methods that use neural radiance fields (NeRFs) show promising results in reproducing the appearance of an object or scene, they do not reconstruct an actual surface.
The volumetric representation of the surface based on densities leads to artifacts when a surface is extracted using Marching Cubes, since during optimization, densities are accumulated along the ray and are not used at a single sample point in isolation.
Instead of this volumetric representation of the surface, we propose to represent the surface using an implicit function (truncated signed distance function).
We show how to incorporate this representation in the NeRF framework, and extend it to use depth measurements from a commodity RGB-D sensor, such as a Kinect.
In addition, we propose a pose and camera refinement technique which improves the overall reconstruction quality.
In contrast to concurrent work on integrating depth priors in NeRF which concentrates on novel view synthesis, our approach is able to reconstruct high-quality, metrical 3D reconstructions.
\end{abstract}

\section{Introduction}

Research on neural networks for scene representations and image synthesis has made impressive progress in recent years~\cite{tewari2020neuralrendering}.
Methods that learn volumetric representations~\cite{mildenhall2020nerf, Lombardi_2019} from color images captured by a smartphone camera can be employed to synthesize near photo-realistic images from novel viewpoints.
While the focus of these methods lies on the reproduction of color images, they are not able to reconstruct metric and clean (noise-free) meshes.
To overcome these limitations, we show that there is a significant advantage in taking additional range measurements from consumer-level depth cameras into account.
Inexpensive depth cameras are broadly accessible and are also built into modern smartphones.
While classical reconstruction methods~\cite{curless,izadi2011kinectfusion,niessner2013hashing} that purely rely on depth measurements struggle with the limitations of physical sensors (noise, limited range, transparent objects, etc.), a neural radiance field-based reconstruction formulation allows to also leverage the dense color information.
Methods like BundleFusion~\cite{dai2017bundlefusion} take advantage of color observations to compute sparse SIFT~\cite{sift} features for re-localization and refinement of camera poses (loop closure).
For the actual geometry reconstruction (volumetric fusion), only the depth maps are taken into account.
Missing depth measurements in these maps, lead to holes and incomplete geometry in the reconstruction.
This limitation is also shared by learned surface reconstruction methods that only rely on the range data~\cite{occnet,sitzmann2019siren}.
In contrast, our method is able to reconstruct geometry in regions where only color information is available.
Specifically, we adapt the neural radiance field (NeRF) formulation of Mildenhall et al.~\cite{mildenhall2020nerf} to learn a truncated signed distance field (TSDF), while still being able to leverage differentiable volumetric integration for color reproduction.
To compensate for noisy initial camera poses which we compute based on the depth measurements, we jointly optimize our scene representation network with the camera poses.
The implicit function represented by the scene representation network allows us to predict signed distance values at arbitrary points in space which is used to extract a mesh using Marching Cubes.
Concurrent work that incorporates depth measurements in NeRF focuses on novel view synthesis~\cite{kangle2021dsnerf,donerf,wei2021nerfingmvs}, and uses the depth prior to restrict the volumetric rendering to near-surface regions~\cite{wei2021nerfingmvs,donerf} or adds an additional constraint on the depth prediction of NeRF~\cite{kangle2021dsnerf}.
NeuS~\cite{wang2021neus} is also a concurrent work on novel view synthesis which uses a signed distance function to represent the geometry, but takes only RGB images as input, and thus fails to reconstruct the geometry of featureless surfaces, like white walls.
In contrast, our method aims for high-quality 3D reconstructions of room-scale scenes using an implicit surface representation and direct SDF-based losses on the input depth maps.
Comparisons to state-of-the-art scene reconstruction methods show that our approach improves the quality of geometry reconstructions both qualitatively and quantitatively.

In summary, we propose an RGB-D based scene reconstruction method that leverages both dense color and depth observations.
It is based on an effective incorporation of depth measurements into the optimization of a neural radiance field using a signed distance-based surface representation to store the scene geometry.
It is able to reconstruct geometry detail that is observed by the color images, but not visible in the depth maps.
In addition, our pose and camera refinement technique is able to compensate for misalignments in the input data, resulting in state-of-the-art reconstruction quality which we demonstrate on synthetic as well as on real data from ScanNet~\cite{dai2017scannet}.

\begin{figure*}[t!]
    \begin{center}
    \includegraphics[width=\linewidth]{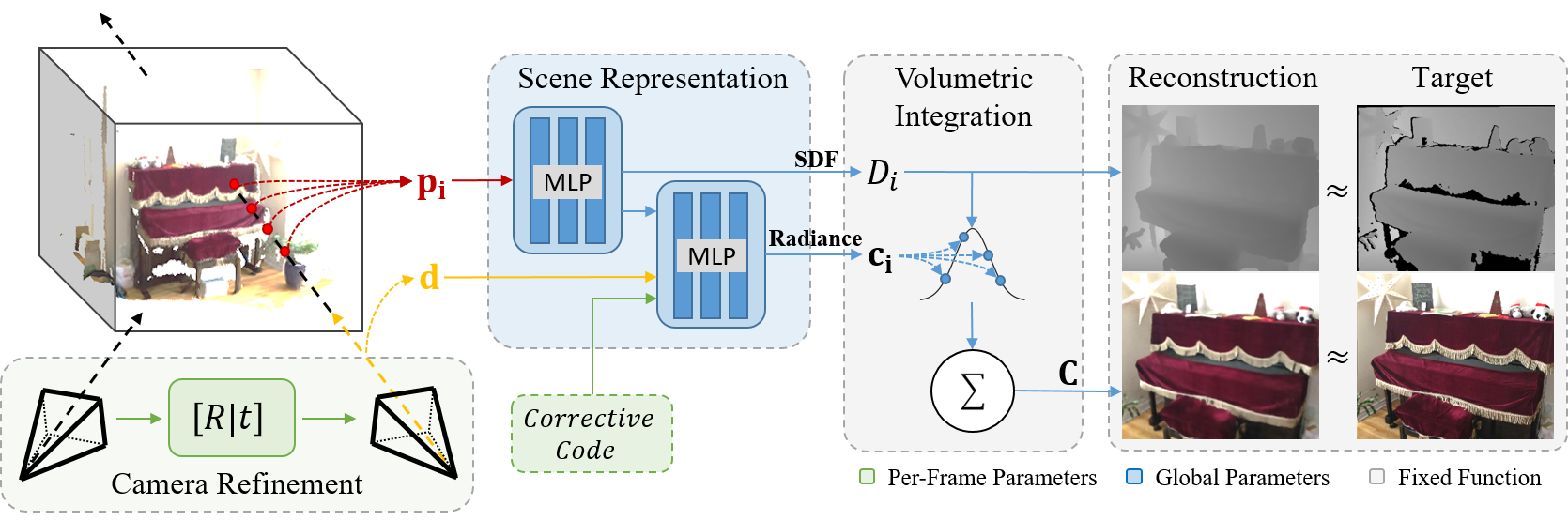}
    \end{center}
   \vspace{-0.5cm}
   \caption{
Differentiable volumetric rendering is used to reconstruct a scene that has been captured using an RGB-D camera.
The scene is represented using multi-layer perceptrons (MLPs), encoding a signed distance value $D_i$ and a viewpoint-dependent radiance value $\textbf{c}_i$ per point $\textbf{p}_i$. 
We perform volumetric rendering by integrating the radiance along a ray, weighing the samples as a function of their signed distance $D_i$ and their visibility. 
We also learn a per-frame latent corrective code to account for exposure or white balance changes throughout the capture, which is passed to the radiance MLP alongside the ray direction $\textbf{d}$.
We optimize the scene representation's MLPs, together with the per-frame corrective codes, the input camera poses, and an image-plane deformation field (not shown) by computing losses for the signed distance $D_i$ of the samples, and the final integrated color $\textbf{C}$ with respect to the input depth and color views.
}
\vspace{-0.1cm}
\label{fig:pipeline}
\end{figure*}

\section{Related Work}

Our approach reconstructs geometry from a sequence of RGB-D frames, leveraging both dense color and depth information.
It is related to classical fusion-based 3D reconstruction methods~\cite{Zollhoefer2018RecoSTAR,curless,newcombe2011kinectfusion,niessner2013hashing,dai2017bundlefusion}, learned 3D reconstruction~\cite{occnet,Peng2020ECCV,chibane20ifnet,dai2021spsg,Weder_2020_CVPR}, as well as to recent coordinate-based scene representation models~\cite{srns,mildenhall2020nerf,tewari2021advances}.

\paragraph{Classical 3D Reconstruction.}
There exists a wide range of methods for RGB and RGB-D based 3D reconstruction that are not based on deep learning.
Reconstructing objects and scenes can be done using passive stereo systems that rely on stereo matching from two or multiple color views~\cite{scharstein2001stereo,goesele2007multiview}, Structure-from-Motion~\cite{Schonberger_2016_CVPR}, or SLAM-based~\cite{engel2013visualodometry,engel14eccv,forster2014svo} methods. 
These approaches may use disjoint representations, like oriented patches~\cite{PMVS}, volumes~\cite{voxel-carving}, or meshes~\cite{hiep2009towards} to reconstruct the scene or object.
Zollhöfer et al.~\cite{Zollhoefer2018RecoSTAR} review the 3D reconstruction methods that rely on range data from RGB-D cameras like the Kinect.
Most of these methods are based on \cite{curless}, where multiple depth measurements are fused using a  signed distance function (SDF) which is stored in a uniform 3D grid.
E.g., KinectFusion~\cite{newcombe2011kinectfusion} combines such representation with real-time tracking to reconstruct objects and small scenes in real-time.
To handle large scenes Nießner et al.~\cite{niessner2013hashing} propose a memory-efficient storage of the SDF grid using spatial hashing.
To handle the loop closure problem when scanning large-scale scenes, bundle adjustment can be used to refine the camera poses \cite{dai2017bundlefusion}.
In addition, several regularization techniques have been proposed to handle outliers during reconstruction~\cite{zach4408983,savinov7780958,dong2018}.

\paragraph{Deep Learning for 3D Reconstruction.}
To reduce artifacts from classical reconstruction methods, a series of methods was proposed that use learned spatial priors to predict depth maps from color images~\cite{laina2016deeper,godard2017unsupervised,fu2018deep}, to learn multi-view stereo using 3D CNNs on voxel grids~\cite{kar2017learning,yao2018mvsnet,sitzmann2019deepvoxels}, or multi-plane images~\cite{flynn2016deepstereo}, to reduce the influence of noisy depth values~\cite{Weder_2020_CVPR}, to complete incomplete scans~\cite{dai2020sgnn,dai2021spsg}, to learn image features for SLAM~\cite{Bloesch_2018_CVPR,Zhi_2019_CVPR,Czarnowski:2020:10.1109/lra.2020.2965415} or feature fusion~\cite{Weder_2021_CVPR,sun2021neucon,bozic2021transformerfusion}, to predict normals~\cite{zhang2018deep}, or to predict objects or parts of a room from single images~\cite{meshrcnn,nie2020total3dunderstanding,dahnert2021panoptic,wang2018pixel2mesh,denninger2020}.
Most recently coordinate-based models have become popular~\cite{tewari2021advances}.
These models use a scene representation that is based on a deep neural network with fully connected layers, i.e., a multi-layer perceptron (MLP)~\cite{tewari2020neuralrendering,tewari2021advances}.
As input the MLP takes a 3D location in the model space and outputs for example, occupancy~\cite{occnet,Peng2020ECCV,chibane20ifnet,pifuSHNMKL19,saito2020pifuhd,DVR,Oechsle2021ICCV}, density~\cite{mildenhall2020nerf}, radiance~\cite{mildenhall2020nerf}, color~\cite{Oechsle2019ICCV}, or the signed distance to the surface~\cite{Park_2019_CVPR,yariv2020multiview,yariv2021volume}.
Scene Representation Networks~\cite{srns} combine such a representation with a learned renderer which is inspired by classical sphere tracing, to reconstruct objects from single RGB images.
Instead, Mildenhall et al.~\cite{mildenhall2020nerf} propose a method that represents a scene as a neural radiance field (NeRF) using a coordinate-based model, and a classical, fixed volumetric rendering formulation~\cite{max}.
Based on this representation, they show impressive novel view synthesis results, while only requiring color input images with corresponding camera poses and intrinsics.
Besides the volumetric image formation, a key component of the NeRF technique is a positional encoding layer, that uses sinusoidal functions to improve the learning properties of the MLP.
In follow-up work, alternatives to the positional encoding were proposed, such as Fourier features~\cite{tancik2020fourfeat} or sinusoidal activation layers~\cite{sitzmann2019siren}.
NeRF has been extended to handle in-the-wild data with different lighting and occluders~\cite{martinbrualla2020nerfw}, dynamic scenes~\cite{li2020neural,park2020nerfies}, avatars~\cite{nerface}, and adapted for generative modeling~\cite{piGAN,Schwarz2020NEURIPS} and image-based rendering~\cite{yu2020pixelnerf,wang2021ibrnet}.
Others have focused on resectioning a camera given a learned NeRF~\cite{yen2020inerf}, and optimizing for the camera poses while learning a NeRF~\cite{wang2021nerf, lin2021barf}.

In our work, we take advantage of the volumetric rendering of NeRF and propose the usage of a hybrid scene representation that consists of an implicit surface representation (SDF) and a volumetric radiance field.
We incorporate depth measurements in this formulation to achieve robust and metric 3D reconstructions.
In addition, we propose a camera refinement scheme to further improve the quality of the reconstruction.
In contrast to NeRF which uses a density based volumetric representation of the scene, our implicit surface representation leads to high quality geometry estimates of entire scenes.

\paragraph{Concurrent Work.}
In concurrent work, Wang et al.~\cite{wang2021neus} present NeuS which uses an implicit surface representation to improve novel view synthesis of NeRF.
Wei et al.~\cite{wei2021nerfingmvs} propose a multi-view stereo approach to estimate dense depth maps which they use to constrain the sampling region when optimizing a NeRF.
Similarly, Neff et al.~\cite{donerf} restrict the volumetric rendering to near surface regions.
Additional constraints on the depth predictions of NeRF were proposed by Deng et al.~\cite{kangle2021dsnerf}.
In contrast to these, our method focuses on accurate 3D reconstructions of room-scale scenes, with explicit incorporation of depth measurements  using an implicit surface representation.

\section{Method}
We propose an optimization-based approach for geometry reconstruction from an RGB-D sequence of a consumer-level camera (e.g., a Microsoft Kinect).
We leverage both the $N$ color frames $\mathcal{I}_i$ as well as the corresponding aligned depth frames $\mathcal{D}_i$ to optimize a coordinate-based scene representation network.
Specifically, our hybrid scene representation consists of an implicit surface representation based on a truncated signed distance function (TSDF) and a volumetric representation for the radiance.
As illustrated in Fig.~\ref{fig:pipeline}, we use differentiable volumetric integration of the radiance values~\cite{max} to compute color images from this representation.
Besides the scene representation network, we optimize for the camera poses and intrinsics.
We initialize the camera poses $\mathcal{T}_i$ using BundleFusion~\cite{dai2017bundlefusion}.
At evaluation time, we use Marching Cubes~\cite{marching_cubes} to extract a triangle mesh from the optimized implicit scene representation.

\subsection{Hybrid Scene Representation}
Our method is built upon a hybrid scene representation which combines an implicit surface representation with a volumetric appearance representation.
Specifically, we implement this representation using a multi-layer perceptron (MLP) which can be evaluated at arbitrary positions $\textbf{p}_i$ in space to compute a truncated signed distance value $D_i$ and view-dependent radiance value $\textbf{c}_i$.
As a conditioning to the MLP, we use a sinusoidal positional encoding $\gamma(\cdot)$~\cite{mildenhall2020nerf} to encode the 3D query point $\textbf{p}_i$ and the viewing direction $\textbf{d}$.

Inspired by the recent success of volumetric integration in neural rendering~\cite{mildenhall2020nerf}, we render color as a weighted sum of radiance values along a ray.
Instead of computing the weights as probabilities of light reflecting at a given sample point based on the density of the medium~\cite{mildenhall2020nerf}, we compute weights directly from signed distance values as the product of two sigmoid functions:
\begin{equation}
    w_i = \sigma\left(\frac{D_i}{tr}\right) \cdot \sigma\left(-\frac{D_i}{tr}\right),
\end{equation}
where $tr$ is the truncation distance.
This bell-shaped function has its peak at the surface, i.e., at the zero-crossing of the signed distance values.
A similar formulation is used in concurrent work~\cite{wang2021neus}, since this function produces unbiased estimates of the signed distance field.
The truncation distance $tr$ directly controls how quickly the weights fall to zero as the distance from the surface increases.
To account for the possibility of multiple intersections, weights of samples beyond the first truncation region are set to zero.
The color along a specific ray is approximated as a weighted sum of the $K$ sampled colors:
\begin{equation}
    \textbf{C} = \frac{1}{\sum_{i=0}^{K-1} w_i} \sum_{i=0}^{K-1} w_i \cdot \textbf{c}_i .
\end{equation}

This scheme gives the highest integration weight to the point on the surface, while points farther away from the surface have lower weights.
Although such an approach is not derived from a physically-based rendering model, as is the case with volumetric integration over density values, it represents an elegant way to render color in a signed distance field in a differentiable manner, and we show that it helps deduce depth values through a photometric loss (see Sec.~\ref{sec:results}).
In particular, this approach allows us to predict hard boundaries between occupied and free space which results in high-quality 3D reconstructions of the surface.
In contrast, density-based models~\cite{mildenhall2020nerf} can introduce semi-transparent matter in front of the actual surface to represent view-dependent effects when integrated along a ray.
This leads to noisy reconstructions and artifacts in free space, as can be seen in Sec.~\ref{sec:results}.

\paragraph{Network Architecture}
Our hybrid scene representation network is composed of two MLPs which represent the shape and radiance, as depicted in Fig.~\ref{fig:pipeline}.
The shape MLP takes the encoding of a queried 3D point $\gamma(\textbf{p})$ as input and outputs the truncated signed distance $D_i$ to the nearest surface.
The task of the second MLP is to produce the surface radiance for a given encoded view direction $\gamma(\textbf{d})$ and an intermediate feature output of the shape MLP.
The view vector conditioning allows our method to deal with view-dependent effects like specular highlights, which would otherwise have to be modeled by deforming the geometry.
Since color data is often subject to varying exposure or white-balance, we learn a per-frame latent corrective code vector as additional input to the radiance MLP~\cite{martinbrualla2020nerfw}.

\vspace{-0.2cm}
\paragraph{Pose and Camera Refinement}
The camera poses $\mathcal{T}_i$, represented with Euler angles and a translation vector for every frame, are initialized with BundleFusion~\cite{dai2017bundlefusion} and refined during the optimization.
Inspired by \cite{color_map}, an additional image-plane deformation field in form of a 6-layer ReLU MLP is added as a residual to the pixel location before unprojecting into a 3D ray to account for possible distortions in the input images or inaccuracies of the intrinsic camera parameters.
Note that this correction field is the same for every frame.
During optimization, camera rays are first shifted with the 2D vector retrieved from the deformation field, before being transformed to world space using the camera pose $\mathcal{T}_i$.

\subsection{Optimization}
We optimize our scene representation network by randomly sampling a batch of $P_b$ pixels from the input dataset of color and depth images.
For each pixel $p$ in the batch, a ray is generated using its corresponding camera pose and $S_p$ sample points are generated on the ray.
Our global objective function $\mathcal{L}(\mathcal{P})$ is minimized w.r.t. the unknown parameters $\mathcal{P}$ (the network parameters $\Theta$ and the camera poses $\mathcal{T}_i$) over all $B$ input batches and is defined as:
\begin{align}
    \mathcal{L}(\mathcal{P}) = \sum_{b=0}^{B-1} \lambda_1 \mathcal{L}_{rgb}^b(\mathcal{P}) + \lambda_2 \mathcal{L}_{fs}^b(\mathcal{P}) + \lambda_3 \mathcal{L}_{tr}^b(\mathcal{P}).
\end{align}
$\mathcal{L}_{rgb}^b(\mathcal{P})$ measures the squared difference between the observed pixel colors $\hat{C}_p$ and predicted pixel colors $C_p$ of the $b$-th batch of rays:
\begin{equation}
    \mathcal{L}_{rgb}^b(\mathcal{P}) = \frac{1}{|P_b|} \sum_{p \in {P_b}} (C_p - \hat{C}_p)^2.
\end{equation} 
$\mathcal{L}_{fs}^b$ is a `free-space' objective, which forces the MLP to predict a value of $tr$ for samples $s \in S_p^{fs}$ which lie between the camera origin and the truncation region of a surface:
\begin{equation}
    \mathcal{L}_{fs}^b(\mathcal{P}) = \frac{1}{|P_b|} \sum_{p \in {P_b}} \frac{1}{|S_p^{fs}|} \sum_{s \in S_p^{fs}}  (D_s - tr)^2.
\end{equation}
For samples within the truncation region ($s \in S_p^{tr}$), we apply $\mathcal{L}_{tr}^b(\mathcal{P})$, the signed distance objective of samples close to the surface:
\begin{equation}
    \mathcal{L}_{tr}^b(\mathcal{P}) = \frac{1}{P_b} \sum_{p \in {P_b}} \frac{1}{|S_p^{tr}|} \sum_{s \in S_p^{tr}} (D_s - \hat{D}_s)^2,
\end{equation}
where $D_s$ is the predicted signed distance of sample $s$, and $\hat{D}_s$ the signed distance observed by the depth sensor, along the optical axis.
In our experiments, we use a truncation distance  $tr=5$~cm, and scale the scene so that the truncation region maps to $[-1,1]$ (positive in front of the surface, negative behind).
The $S_p$ sample points on the ray are generated in two steps.
In the first step $S'_c$ sample points are generated on the ray using stratified sampling.
Evaluating the MLP on these $S'_c$ sample points allows us to get a coarse estimate for the ray depth by explicitly searching for the zero-crossing in the predicted signed distance values.
In the second step, another $S'_f$ sample points are generated around the zero-crossing and a second forward pass of the MLP is performed with these additional samples.
The output of the MLP is concatenated to the output from the first step and color is integrated using all $S'_c + S'_f$ samples, before computing the objective loss.
It is important that the sampling rate in the first step is high enough to produce samples within the truncation region of the signed distance field, otherwise the zero-crossing may be missed.
We implement our method in Tensorflow using the ADAM optimizer~\cite{adam} with a learning rate of $5\times10^{-4}$ and set the loss weights to $\lambda_1 = 0.1$, $\lambda_2 = 10$ and $\lambda_3 = 6\times10^3$.
We run all of our experiments for $2\times10^{5}$ iterations, where in each iteration we compute the gradient w.r.t. $|P_b|=1024$ randomly chosen rays.
We set the number of $S'_f$ samples to $16$. $S'_c$ is chosen such that there is on average one sample for every $1.5$ cm of the ray length.
The ray length itself needs to be greater than the largest distance in the scene that is to be reconstructed and ranges from $4$ to $8$ meters in our scenes.

\section{Results}
\label{sec:results}

In the following, we evaluate our method on real, as well as on synthetic data.
For the shown results, we use Marching Cubes~\cite{marching_cubes} with a spatial resolution of $1$~cm to extract a mesh from the reconstructed signed distance function.

\begin{table}
	\resizebox{\linewidth}{!}{%
    \centering
    \begin{tabular}{lcccc}
        \toprule
        \textbf{Method}  & \textbf{C-$\ell_1$} $\downarrow$  & \textbf{IoU} $\uparrow$ & \textbf{NC} $\uparrow$  & \textbf{F-score} $\uparrow$ \\
        \midrule
        BundleFusion     & 0.062                             & 0.594                   & 0.892                   & 0.805	                   \\
        RoutedFusion     & 0.057                             & 0.615                   & 0.864                   & 0.838                       \\
        COLMAP + Poisson & 0.057                             & 0.619                   & 0.901                   & 0.839                       \\
        Conv. Occ. Nets  & 0.077                             & 0.461                   & 0.849                   & 0.643                       \\
        SIREN            & 0.060                             & 0.603                   & 0.893                   & 0.816                       \\
        NeRF + Depth     & 0.065                             & 0.550                   & 0.768                   & 0.782                       \\
        \midrule
        Ours (w/o pose)  & 0.049                             & 0.655                   & 0.908                   & 0.868                       \\
        Ours             & \textbf{0.044}                    & \textbf{0.747}          & \textbf{0.918}          & \textbf{0.924}              \\
        \bottomrule
    \end{tabular}
    }
    \caption{Reconstruction results on a dataset of 10 synthetic scenes. The Chamfer $\ell_1$ distance, normal consistency and the F\mbox{-}score~\cite{10.1145/3072959.3073599} are computed between point clouds sampled with a density of 1 point per cm\textsuperscript{2}, using a threshold of 5~cm for the F-score. We voxelize the mesh to compute the intersection-over-union (IoU) between the predictions and ground truth.}
\vspace{-0.2cm}
\label{tab:quantitative}
\end{table}

\begin{figure*}
\begin{center}
\includegraphics[width=0.975\linewidth]{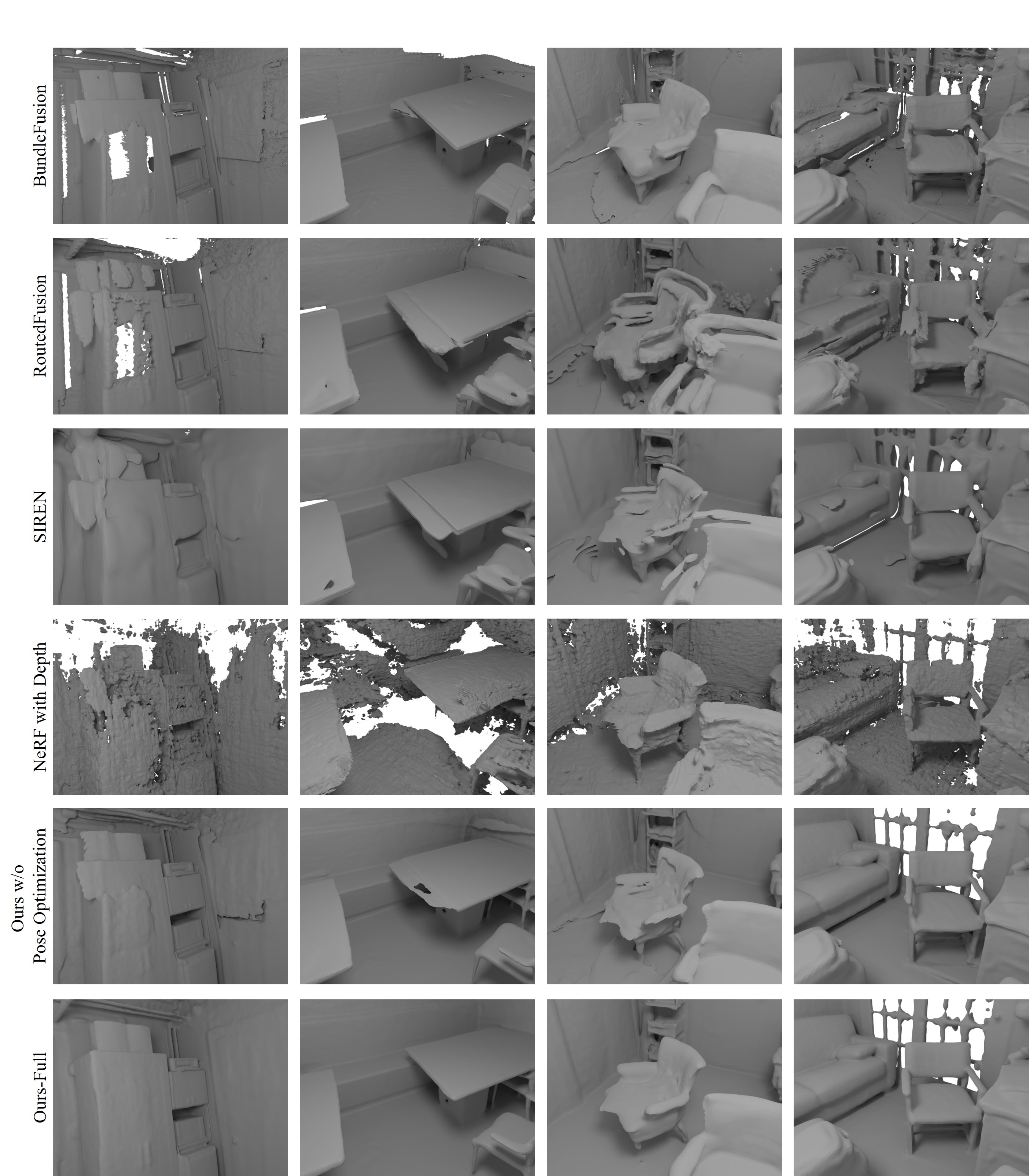}
\end{center}
   \caption{We compare our model without pose optimization and our full model with both the pose optimization and image-plane deformation field to BundleFusion, RoutedFusion, SIREN and a NeRF optimized with depth supervision in scenes 2, 5, 12, and 50 of the ScanNet dataset. Our model without pose optimization recovers smoother meshes than the density-based NeRF model, but still suffers from misalignment artifacts. These are solved by our full model to recover a clean reconstruction.}
\label{fig:bundlefusion_vs_ours}
\end{figure*}

\paragraph{Results on real data.}
We test our method on the ScanNet dataset~\cite{dai2017scannet} which provides RGB-D sequences of room-scale scenes.
The data has been captured with a StructureIO camera which provides quality similar to that of a Kinect v1.
The depth measurements are noisy and often miss structures like chair legs or other thin geometry.
To this end our method proposes the additional usage of a dense color reconstruction loss, since regions that are missed by the range sensor are often captured by the color camera.
To compensate for the exposure and white balancing of the used camera, our approach learns a per-frame latent code as proposed in~\cite{martinbrualla2020nerfw}.
In Fig.~\ref{fig:bundlefusion_vs_ours}, we compare our method to the original ScanNet BundleFusion reconstructions which often suffer from severe camera pose misalignment.
Our approach jointly optimizes for the scene representation network as well as the camera poses, leading to substantially reduced misalignment artifacts in the reconstructed geometry.
%

\begin{figure*}[htb!]
\begin{center}
\vspace{-0.1cm}
\includegraphics[width=\linewidth]{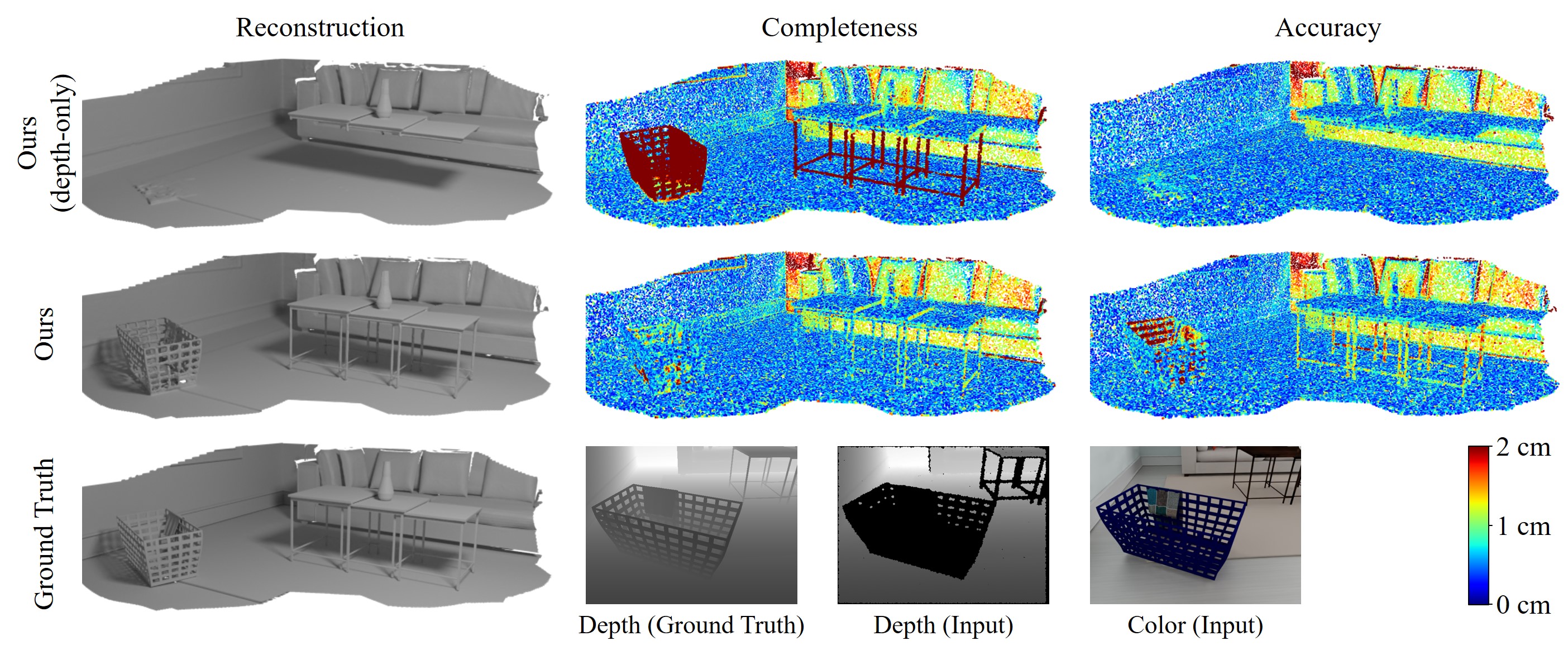}
\end{center}
\vspace{-0.5cm}
   \caption{Accuracy shows how close ground truth points are to predicted points, while completeness shows how close predicted points are to ground truth points. Geometry reconstructed purely through the photometric loss has slightly lower accuracy than geometry for which depth observations were also available. Furthermore, the accuracy and completeness drop in distant areas, which had less multi-view constraints and more noise in the depth measurements.}

\vspace{-0.2cm}
\label{fig:completeness_accuracy}
\end{figure*}

\paragraph{Quantitative evaluation.}
We perform a quantitative evaluation of our method on a dataset of 10 synthetic scenes for which the ground truth geometry and camera trajectory are known.
Note that the ground truth camera trajectory is only used for the rendering and evaluation, and not for the reconstruction.
For each frame, we render a photo-realistic image using Blender~\cite{blender, denninger2019blenderproc}.
We apply noise and artifacts, similar to those of a real depth sensor~\cite{sim_kinect_noise, Barron:etal:2013A, Bohg:etal:2014,handa:etal:2014}.
On this data, we compare our technique to several state-of-the-art methods that use either depth input only, or both color and depth data to reconstruct geometry (see Tab.~\ref{tab:quantitative}).

\emph{BundleFusion.}
BundleFusion~\cite{dai2017bundlefusion} uses the color and depth input to reconstruct the scene. It is a classical depth fusion approach~\cite{curless} which compensates misalignments using a global bundle adjustment approach. 

\emph{RoutedFusion.}
RoutedFusion~\cite{Weder_2020_CVPR} uses a routing network which filters sensor-specific noise and outliers from depth maps and computes pixel-wise confidence values, which are used by a fusion network to produce the final SDF.
%
It takes the depth maps and camera poses as input.

\emph{COLMAP with screened Poisson surface reconstruction.}
We obtain camera poses using COLMAP~\cite{Schonberger_2016_CVPR, schoenberger2016vote, schoenberger2016mvs} and use these to back-project depth maps into world space. 
We obtain a mesh by applying screened Poisson surface reconstruction~\cite{10.1145/2487228.2487237} on the resulting point cloud.

\emph{Convolutional Occupancy Networks.}
We accumulate the point clouds from the depth maps using BundleFusion poses and evaluate the pre-trained convolutional occupancy networks model~\cite{Peng2020ECCV} provided by the authors (which has been used on similar data~\cite{Matterport3D}).

\emph{SIREN.}
We optimize a SIREN~\cite{sitzmann2019siren} per scene using the back-projected point cloud data.
The ICL-NUIM~\cite{handa:etal:2014} scene on which the method was originally tested, is also included in our synthetic dataset.

\emph{NeRF with an additional depth loss.}
NeRF~\cite{mildenhall2020nerf} proposes using the expected ray termination distance as a way to visualize the depth of the scene. 
In our baseline, we add an additional loss to NeRF, where this depth value is compared to the input depth using an L2 loss.
Note that this baseline still uses NeRF's density field to represent geometry.

As can be seen in Tab.~\ref{tab:quantitative}, our approach with camera refinement results in the lowest Chamfer distance, and the highest IoU, normal consistency (mean of the dot product of the ground truth and predicted normals), and F-score~\cite{10.1145/3072959.3073599}.
Especially, the comparison to the density-based NeRF with an additional depth constraint shows the benefit of our proposed hybrid scene representation.

\vspace{-0.2cm}
\paragraph{Ablation studies.}
We conduct ablation studies to justify our choice of network architecture and training parameters.
In Fig.~\ref{fig:bundlefusion_vs_ours}, we show the difference between a volumetric representation (density field, `NeRF with Depth') to an implicit surface representation (signed distance field, `Ours-Full') on real data from ScanNet~\cite{dai2017scannet}.
While representing scenes with a density field works great for color integration, extracting the geometry itself is a challenging problem.
Although small variations in density may not affect the integrated color much, they cause visible noise in the extracted geometry and produce floating artifacts in free space.
These artifacts can be reduced by choosing a different iso-level for geometry extraction with Marching Cubes, but this leads to less complete reconstructions.
In contrast, a signed distance field models a smooth boundary between occupied and free space, and we show that it can be faithfully represented by an MLP.
However, the reconstruction quality is still limited by the provided camera poses, as can be seen in Fig.~\ref{fig:bundlefusion_vs_ours} (e.g., the cabinet in the left column).
Optimizing for pose corrections further improves the quality of our reconstructions.

\begin{figure*}[htb!]
\begin{center}
\includegraphics[width=\linewidth]{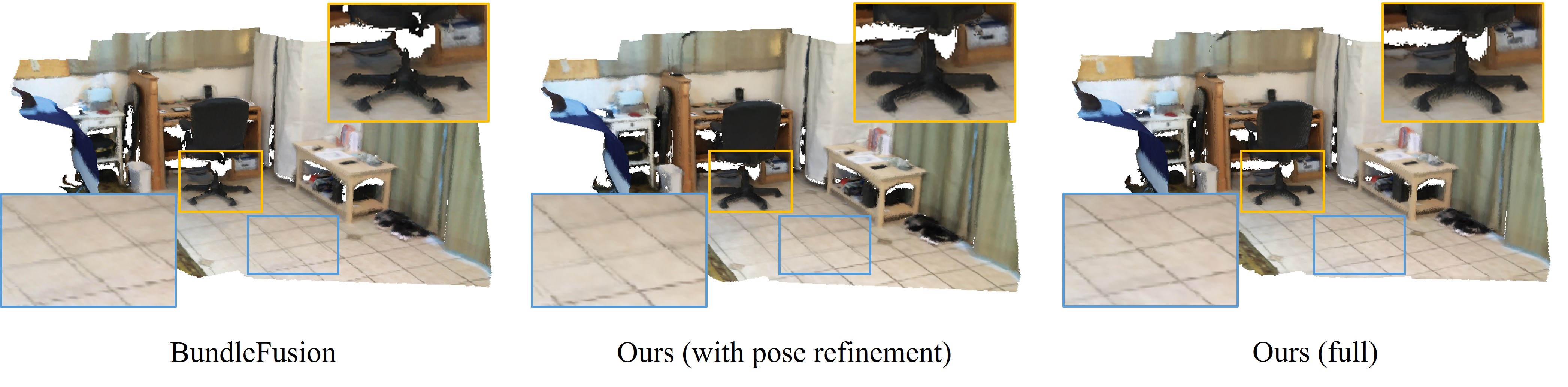}
\end{center}
\vspace{-0.7cm}
   \caption{Our method improves the camera alignment over the baseline, as visible in the tiles of the floor. The additional image-plane distortion correction results in straight and aligned edges in the reconstruction.}
\label{fig:computer_chair_color}
\vspace{-0.15cm}
\end{figure*}

\vspace{-0.1cm}
\paragraph{Effect of the photometric term.}
A fundamental component of our method is the use of a photometric term to infer depth values which are missing from camera measurements.
We analyze the effect of this term on the synthetic scene in Fig.~\ref{fig:completeness_accuracy}, where we simulate missing geometry of the table legs and the meshed basket.
In the figure, we visualize the completeness and accuracy.
In contrast to a model without the photometric term, our method is still able to reconstruct the missing geometry leveraging the RGB observations.

For our full approach, we also separately evaluate the reconstruction quality of geometry where depth measurements were available and where they were missing.
Regions that relied only on color have a somewhat worse average accuracy of $11$~mm, compared to $8$~mm for regions that had access to depth measurements.
We refer the reader to the supplemental material for more details and a qualitative comparison on real data.

\begin{table}
	\resizebox{\linewidth}{!}{%
    \centering
    \begin{tabular}{lcccc}
        \toprule
        \textbf{Method}  & \textbf{Pos. error (meters)} $\downarrow$  & \textbf{Rot. error (degrees)} $\downarrow$   \\
        \midrule
        BundleFusion     & 0.033                                            & 0.571                                               \\
        COLMAP           & 0.038                                            & 0.692                                               \\
        \midrule
        Ours             & \textbf{0.021}                                   & \textbf{0.144}                                      \\
        \bottomrule
    \end{tabular}
    }
    \caption{Based on our synthetic dataset, we evaluate the average positional and rotational errors of the estimated camera poses. Our method is able to further increase the pose estimation accuracy compared to its BundleFusion initialization.}
\vspace{-0.1cm}
\label{tab:quantitative_poses}
\end{table}

\paragraph{Effect of pose refinement.}
We show that initial camera pose estimates can be further improved by jointly optimizing for the rotation and translation parameters of the cameras which are initialized with BundleFusion~\cite{dai2017bundlefusion}.
We quantitatively evaluate this on all scenes in our synthetic dataset.
An aggregate of the positional and rotational errors of different methods is presented in Tab.~\ref{tab:quantitative_poses}.
A detailed per-scene breakdown is given in the supplemental material.
In Tab.~\ref{tab:quantitative} and Fig.~\ref{fig:bundlefusion_vs_ours}, we show that optimizing camera poses reduces geometry misalignment artifacts and improves the overall reconstruction, both quantitatively and qualitatively.

\paragraph{Effect of the image-plane deformation field.}
To evaluate the effect of the pixel-space deformation field, we initialize the camera with an incorrect focal length and optimize our model with and without the deformation field.
Tab.~\ref{tab:quantitative_intrinsics} shows that the deformation field mitigates this inaccuracy in the camera's intrinsic parameters which leads to significantly better reconstruction results compared to the model that does not use the deformation field.
Fig.~\ref{fig:computer_chair_color} showcases the effects of our camera pose and image-plane deformation field~\cite{dai2017bundlefusion}.
Blurry frames and sparse features lead to systematic camera pose errors in BundleFusion.
Our method improves these camera poses and the camera distortion model, and, thus, is able to better align scene features, resulting in higher reconstruction quality.
\paragraph{Limitations and future work.}
Similar to other methods that are based on a scene representation which uses a scene-specific MLP, our method runs offline (around $9$~hours for $2\times10^5$ iterations using an NVIDIA RTX 3090).
Recent methods that utilize voxel grids to optimize a radiance field~\cite{sun2021direct,yu2021plenoxels} have shown significantly faster convergence compared to earlier MLP-based methods and we believe that they would also be good candidates for improving our method.
Nonetheless, our proposed method offers a high-quality scene reconstruction which outperforms online fusion approaches.
Another limitation is the global MLP which stores the entire scene information which comes at the cost of missing high-frequency local detail in very large scenes.
Approaches like IF-Nets~\cite{chibane20ifnet} or Convolutional Occupancy Networks~\cite{Peng2020ECCV} benefit from locally-conditioned MLPs and can be integrated in future work.
Finally, our method was designed to handle only opaque surfaces.

\begin{table}[t]
    \resizebox{\linewidth}{!}{%
    \centering
    \begin{tabular}{lcccc}
        \toprule
        \textbf{Method}  & \textbf{C-$\ell_1$} $\downarrow$  & \textbf{IoU} $\uparrow$ & \textbf{NC} $\uparrow$  & \textbf{F-score} $\uparrow$ \\
        \midrule
        Ours (w/o IPDF)  & 0.061                             & 0.266                   & 0.886                   & 0.406                       \\
        Ours (w/  IPDF)  & \textbf{0.031}                    & \textbf{0.609}          & \textbf{0.911}          & \textbf{0.904}              \\
        \bottomrule
    \end{tabular}
    }
    \vspace{-0.1cm}
    \caption{Ablation of the image-plane deformation field (IPDF) which  compensates image space distortions and incorrect intrinsic parameters. The experiment is based on a synthetic scene, where we assume an incorrect focal length of $570$ instead of $554.26$ (GT).}
\label{tab:quantitative_intrinsics}
\end{table}
\section{Conclusion}

We have presented a new method for 3D surface reconstruction from RGB-D sequences by introducing a hybrid scene representation that is based on an implicit surface function and a volumetric representation of radiance.
This allows us to efficiently incorporate depth observations, while still benefiting from the differentiable volumetric rendering of the original neural radiance field formulation.
As a result, we obtain high-quality surface reconstructions, outperforming  traditional and learned RGB-D fusion methods.
Overall, we believe our work is a stepping stone towards leveraging the success of implicit, differentiable representations for 3D surface reconstruction.

\section*{Acknowledgements}
{
This work was supported by a Google Gift Grant, a TUM-IAS Rudolf M{\"o}{\ss}bauer Fellowship, an NVidia Professorship Award, the ERC Starting Grant Scan2CAD (804724), and the German Research Foundation (DFG) Grant Making Machine Learning on Static and Dynamic 3D Data Practical.
We would also like to thank Angela Dai for the video voice-over.
}

{\small
\bibliographystyle{ieee_fullname}
\bibliography{egbib}
}

\begin{appendix}
\clearpage
\newpage
\twocolumn[{%
\renewcommand\twocolumn[1][]{#1}%
\maketitle
\begin{center}
	\includegraphics[width=\linewidth]{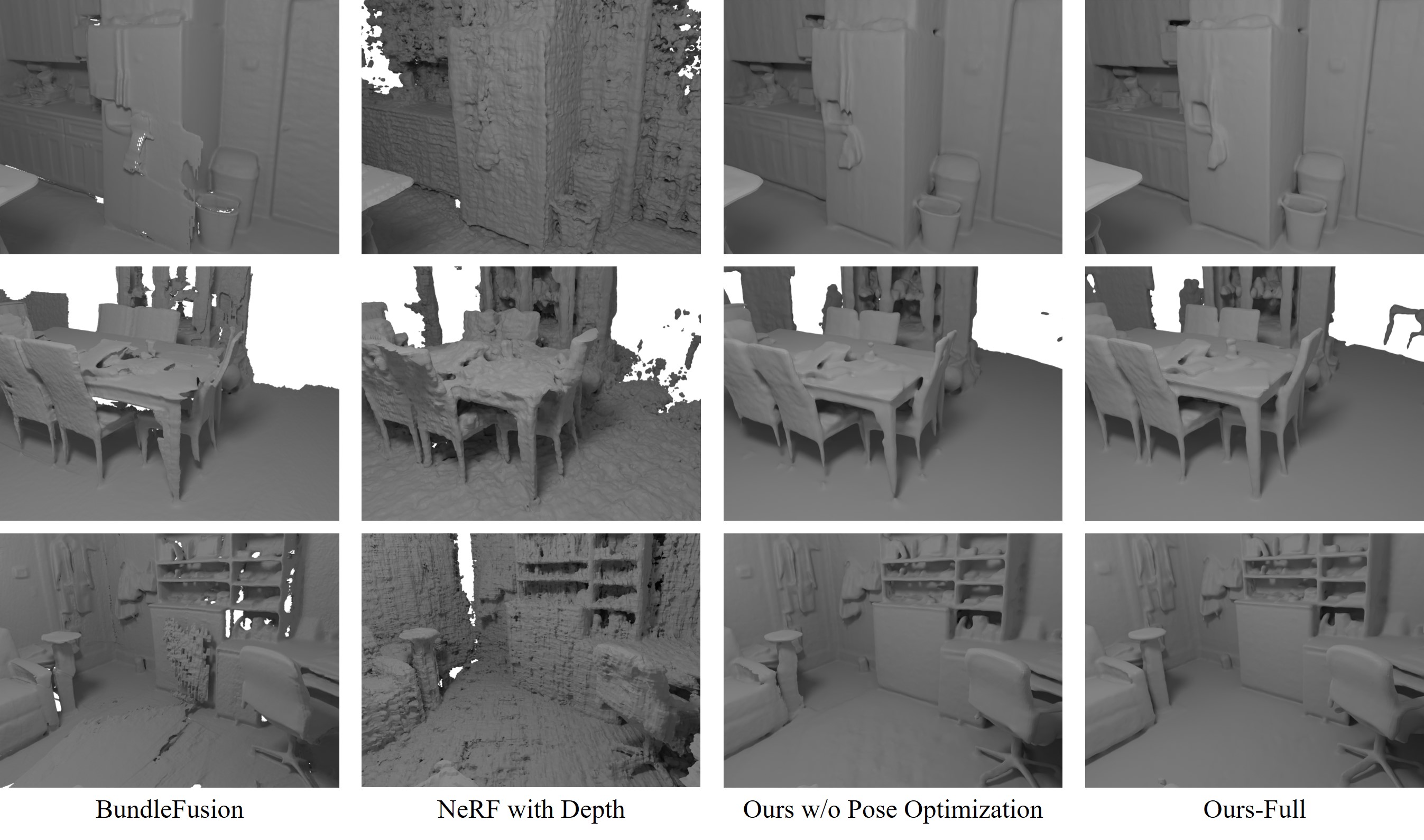}
    \captionof{figure}{
    Our method obtains a high-quality 3D reconstruction from an RGB-D input sequence by training a multi-layer perceptron. In comparison to state-of-the-art methods like BundleFusion~\cite{dai2017bundlefusion} or the theoretical NeRF~\cite{mildenhall2020nerf} with additional depth constraints, our approach results in cleaner and more complete reconstructions. As can be seen, the pose optimization of our approach is key to resolving misalignment artifacts.
}
\vspace{0.5cm}
\label{fig:appendix_teaser}
\end{center}
}]

\section*{APPENDIX}
In this appendix we show a per-scene breakdown of the quantitative evaluation from Tab.~\ref{tab:quantitative}, an ablation study on additional scenes from the ScanNet dataset (see Fig.~\ref{fig:appendix_teaser}), as well as further ablation studies on synthetic data.
For the purpose of reproducibility, we also provide further details on the parameters that were used for optimization in each of the scenes.

\section{Implementation Details}
We implement our method in TensorFlow v2.4.1 using the ADAM~\cite{adam} optimizer with a learning rate of $5\times10^{-4}$ and an exponential learning rate decay of $10^{-1}$ over $2.5\times10^5$ iterations.
In each iteration, we compute a gradient w.r.t. $|P_b|=1024$ randomly chosen rays.
We set the number of $S'_f$ samples to 16. $S'_c$ is chosen so that there is on average one sample for every 1.5 cm of the ray length.
Tab.~\ref{tab:num_samples} gives an overview of ray length and number of samples for each of the experiments.
Internally, we translate and scale each scene so that it lies within a $[-1, 1]^3$ cube.
Depending on scene size, our method takes between $9$ and $13$ hours to converge on a single NVIDIA RTX 3090 (see Sec.~\ref{sec:timings}).
We set the loss weights to $\lambda_1 = 0.1$, $\lambda_2 = 10$ and $\lambda_3 = 6\times10^3$.
We use $8$ bands for the positional encoding of the point coordinates and $4$ bands to encode the view direction vector.
To account for distortions or inaccuracies of the intrinsic parameters, a 2D deformation field of the camera pixel space in form of a 6-layer MLP, with a width of 128, is used.

\begin{table}[h]
	
    \centering
    \begin{tabular}{ L{2.75cm} C{0.55cm} C{1.8cm} C{1.5cm} }
\toprule
Scene       &  $S'_c$  & $ray$ $length$ (m) & $\# frames$\\
\midrule
Scene 0     & 512      & 8                  & 1394      \\
Scene 2     & 256      & 4                  & 1299      \\
Scene 5     & 256      & 4                  & 1159      \\
Scene 12    & 320      & 5                  & 1335      \\
Scene 24    & 512      & 8                  & 849       \\
Scene 50    & 256      & 4                  & 1163      \\
Scene 54    & 256      & 4                  & 1250      \\
\midrule
Breakfast room    & 320      & 5                  & 1167      \\
Green room        & 512      & 8                  & 1442      \\
Grey-white room   & 512      & 8                  & 1493      \\
ICL living room   & 320      & 5                  & 1510      \\
Kitchen 1         & 512      & 8                  & 1517      \\
Kitchen 2         & 640      & 10                 & 1221      \\
Morning apartment & 256      & 4                  & 920       \\
Staircase         & 512      & 8                  & 1149      \\
Thin geometry     & 256      & 4                  & 395       \\
White room        & 512      & 8                  & 1676      \\
\bottomrule
\end{tabular}
\vspace{-0.15cm}
\caption{We list the number of samples $S'_c$ and the ray length in meters that were used to reconstruct each of the ScanNet scenes and the synthetic scenes. Note that these settings are dependent on the scene size.}
\label{tab:num_samples}
\end{table}

\section{Per-scene Quantitative Evaluations}

In Tab.~\ref{tab:quantitative_per_scene_1} and Tab.~\ref{tab:quantitative_per_scene_2} we present a per-scene breakdown of the quantitative analysis from the main paper (see Sec.~4, Tab.~1 and Tab.~2 in the main paper).
The corresponding qualitative results are shown in Fig.~\ref{fig:synth_comparison_1} and Fig.~\ref{fig:synth_comparison_2}.

\paragraph{Reconstruction Evaluation.}
The goal of our method is to reconstruct a scene from color and depth data, i.e., we do not aim for scene completion.
To evaluate the reconstruction quality, we evaluate the quality of reconstructions w.r.t. Chamfer distance~(\mbox{C-$\ell_1$}), intersection-over-union (IoU), normal consistency (NC) based on cosine similarity, and F-score.
These metrics are computed on surfaces which were visible in the color and depth streams (geometry within the viewing frusta of the input images).
Specifically, we subdivide all meshes to have a maximum edge length of below $1.5$~cm and use the ground truth trajectory to detect vertices which are visible in at least one camera.
Triangles which have no visible vertices, either due to not being in any of the viewing frusta or due to being occluded by other geometry, are culled.
This is necessary to avoid computing the error in regions such as occluded geometry in the synthetic ground truth mesh or in regions where the network output is unpredictable because the region was never seen at training time.
The culled geometry is sampled with a density of 1 point per cm$^2$ and the error metrics are evaluated on the sampled point clouds.
To evaluate the IoU, we voxelize the reconstruction using voxels with an edge length of $5$~cm.
The F-score is also computed using a $5$~cm threshold.

\paragraph{Synthetic Dataset.}
Our synthetic dataset which we use for numeric evaluation purposes consists of 10 scenes published under either the CC-BY or CC-0 license (see Tab.~\ref{tab:scene_urls}).
We define a trajectory by a Catmull-Rom spline interpolation~\cite{catmull_rom} on several manually chosen control points.
We use BlenderProc~\cite{denninger2019blenderproc} to render color and depth images  for each camera pose in the interpolated trajectory.
Noise is applied to the depth maps to simulate sensor noise of a real depth sensor~\cite{sim_kinect_noise, Barron:etal:2013A, Bohg:etal:2014,handa:etal:2014}.
For the ICL scene~\cite{handa:etal:2014}, we use the color and noisy depth provided by the authors and do not render our own images.
The scenes in the dataset have various sizes, complexity and materials like highly specular surfaces or mirrors.
BundleFusion~\cite{dai2017bundlefusion} is used to get an initial estimate of the camera trajectory.
This estimated trajectory is used by all methods other than COLMAP to allow a fair comparison.

\newcommand {\datalinks}[2]{\href{#1}{#2}}
\newcommand {\datalink}[1]{\href{#1}{#1}}
\begin{table}[]
	\resizebox{\linewidth}{!}{
    \centering
    \begin{tabular}{ lll }
        \toprule
        Scene       &  URL  & License\\
        \midrule
        ScanNet     & \datalink{http://www.scan-net.org/}      & MIT \\
        \midrule
        Breakfast room    & \datalink{https://blendswap.com/blend/13363} & CC-BY \\
        Green room        & \datalink{https://blendswap.com/blend/8381}  & CC-BY \\
        Grey-white room   & \datalink{https://blendswap.com/blend/13552} & CC-BY \\
        ICL living room   & \datalink{https://www.doc.ic.ac.uk/~ahanda/VaFRIC/iclnuim.html} & CC-BY \\
        Kitchen 1         & \datalink{https://blendswap.com/blend/5156} & CC-BY  \\
        Kitchen 2         & \datalink{https://blendswap.com/blend/11801}  & CC-0 \\
        Morning apart.    & \datalink{https://blendswap.com/blend/10350} & CC-0  \\
        Staircase         & \datalink{https://blendswap.com/blend/14449} & CC-BY \\
        Thin geometry     & \datalink{https://blendswap.com/blend/8381}  & CC-BY \\
        White room        & \datalink{https://blendswap.com/blend/5014}  & CC-BY \\
        \bottomrule
    \end{tabular}
    }
    \vspace{-0.15cm}
    \caption{Source and license information of the used data.}
    \label{tab:scene_urls}
    
\end{table}

\vspace{0.4cm}
\begin{figure}[htb!]
\begin{center}
\includegraphics[width=\linewidth]{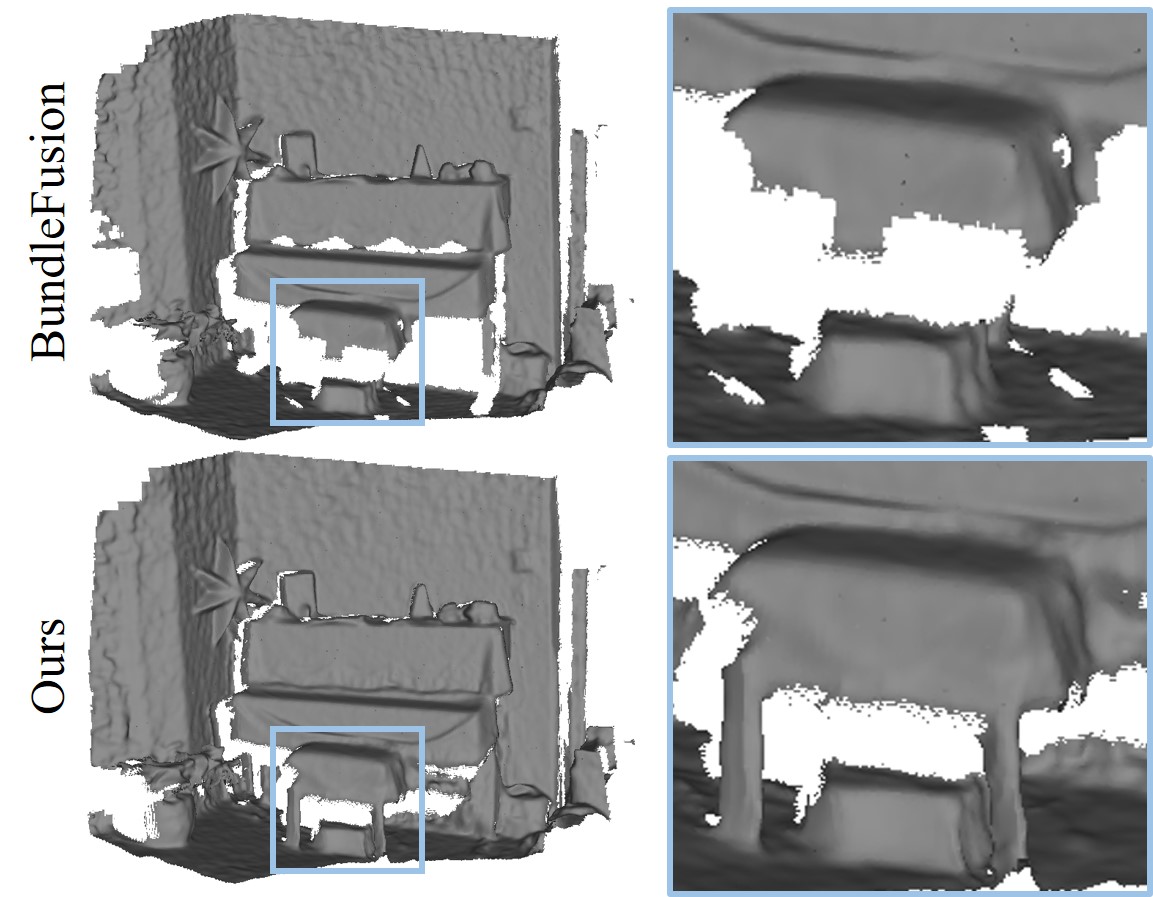}
\end{center}
\vspace{-0.2cm}
   \caption{The photometric energy term encourages correct depth prediction in areas where the depth sensor did not capture any depth measurements.}
\label{fig:piano_stool}
\end{figure}

\section{Ablation Studies}

In this section, we present additional details for the ablation studies described in the main paper, and show further studies to test the robustness and the limitations of our method.
In Fig.~\ref{fig:appendix_teaser}, the additional results on real data demonstrate the advantages of the signed distance field and our camera refinement.

\begin{table}[htb!]
    \resizebox{\linewidth}{!}{%
    \centering
    \begin{tabular}{lcccc}
        \toprule
        \textbf{Method}   & \textbf{C-$\ell_1$} $\downarrow$  & \textbf{IoU} $\uparrow$ & \textbf{NC} $\uparrow$  & \textbf{F-score} $\uparrow$ \\
        \midrule
        Ours (depth-only) & 0.017                             & 0.791                   & \textbf{0.910}          & 0.944                       \\
        Ours (full)       & \textbf{0.009}                    & \textbf{0.865}          & \textbf{0.910}          & \textbf{0.995}              \\
        \bottomrule
    \end{tabular}
    }
    \caption{Detailed reconstruction results for Fig.~4 from the main paper. Our method reconstructs geometry visible only in color images, leading to significantly better reconstruction results in scenes with geometry which is not captured by the depth sensor.}
\label{tab:photometric_details}
\end{table}

\subsection{Effect of the Photometric Energy Term}

In Tab.~\ref{tab:photometric_details}, we list the quantitative evaluation of the experiment on the effectiveness of the photometric energy term from Fig.~4 in the main paper.
Fig.~\ref{fig:piano_stool} shows the effect of the term on a real scene from the ScanNet dataset.
The legs of the piano stool were not visible in any of the depth maps.
Nevertheless, our method is able to reconstruct them by making use of the corresponding color data.

\begin{figure}[b]
\begin{center}
\includegraphics[width=\linewidth]{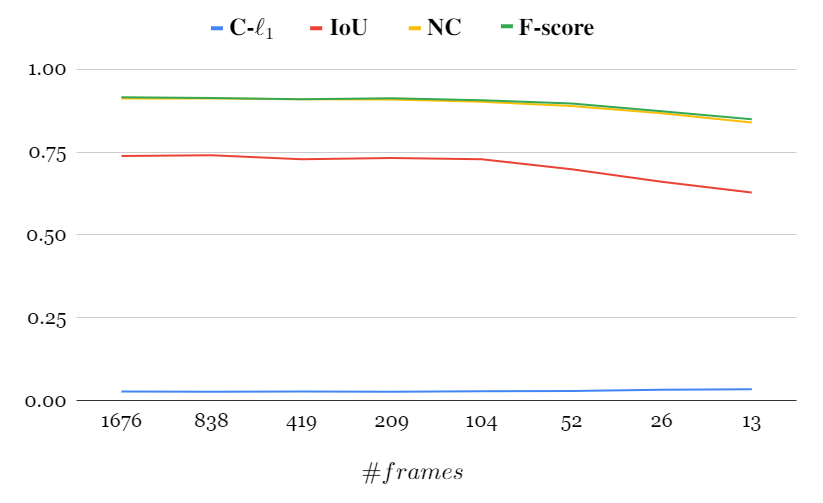}
\end{center}
\vspace{-0.75cm}
   \caption{We test the robustness of our method by removing frames from the dataset used for optimization. Our method achieves good reconstruction results using as few as 13 frames.}
\label{fig:frame_skip}
\end{figure}

\subsection{Number of Input Frames}

The reconstruction quality of any reconstruction method is dependent on the number of input frames.
We evaluate our method on the `whiteroom' synthetic scene through multiple experiments in which we remove different numbers of frames in the dataset used for optimization.
Reconstruction results are presented in Fig.~\ref{fig:frame_skip}.
Note that for these experiments we use the camera poses initialized with BundleFusion which uses all $1676$ depth frames.

\subsection{Robustness to Noisy Pose Initialization}
\begin{figure}[htb!]
\begin{center}
\includegraphics[width=\linewidth]{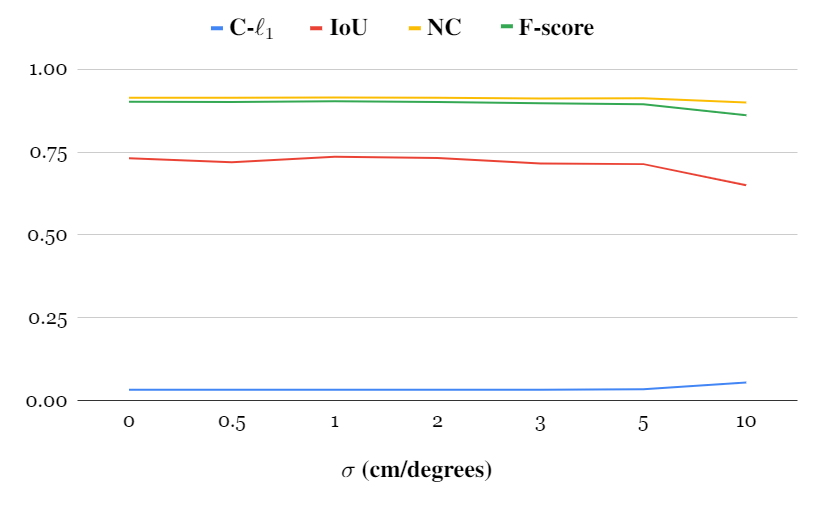}
\end{center}
\vspace{-0.5cm}
   \caption{We test the robustness of our reconstructions to noise in the initial camera position and direction. Our method achieves good results even in the presence of significant noise. At $\sigma$~=~$10$~cm, some of the cameras intersect geometry, degrading the reconstruction quality.}
\label{fig:pose_robustness}
\end{figure}

\begin{figure}[htb!]
\begin{center}
\includegraphics[width=\linewidth]{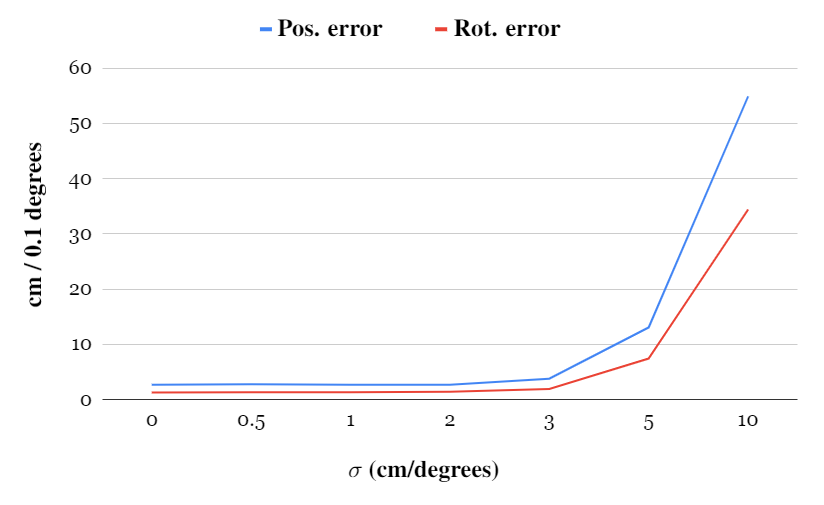}
\end{center}
\vspace{-0.5cm}
   \caption{We test the robustness of our pose refinement to noise in the initial camera position and direction. The rotation error has been scaled by a factor of $10$ for better visibility. Our method is able to correct poses even in the presence of significant noise. At $\sigma$~=~$10$~cm, some of the cameras start intersecting geometry, making refinement impossible.}
\label{fig:pose_robustness_cam_error}
\end{figure}

To analyze the robustness of our method w.r.t. presence of inaccuracies in camera alignment, we apply Gaussian noise to every camera's position and direction in the `whiteroom' scene.
In Fig.~\ref{fig:pose_robustness} we present reconstruction results for poses of increasing inaccuracy.
We separately show the pose errors of the refined cameras in Fig.~\ref{fig:pose_robustness_cam_error}.
On the reconstruction metrics, our method is robust to camera position and orientation errors of up to $5$~cm and $5^{\circ}$ respectively.
The pose refinement is robust up to a noise level of $3$~cm and $3^{\circ}$.
At noise levels with a standard deviation of $10$~cm and higher, some cameras are initially positioned inside geometry, preventing our method from refining their position and leading to large errors in geometry reconstruction.

\vspace{0.2cm}

\begin{table}[b]
    \centering
    \resizebox{\linewidth}{!}{%
    \begin{tabular}{ccccc}
        \toprule
        \textbf{Truncation (cm)}  & \textbf{C-$\ell_1$} $\downarrow$  & \textbf{IoU} $\uparrow$ & \textbf{NC} $\uparrow$  & \textbf{F-score} $\uparrow$ \\
        \midrule
        2                  & 0.053          & 0.671          & 0.855          & 0.862                       \\
        3                  & 0.023          & 0.766          & 0.901          & 0.930              \\
        5                  & \textbf{0.021}          & \textbf{0.786}          & \textbf{0.912}          & \textbf{0.933}                       \\
        10                 & 0.024          & 0.742          & 0.908          & 0.912              \\
        \bottomrule
    \end{tabular}
    }
    \caption{Impact of the truncation region width on reconstruction quality.}
\label{tab:truncation}
\end{table}

\subsection{Batch Size}

\begin{figure}[t]
\begin{center}
\includegraphics[width=\linewidth]{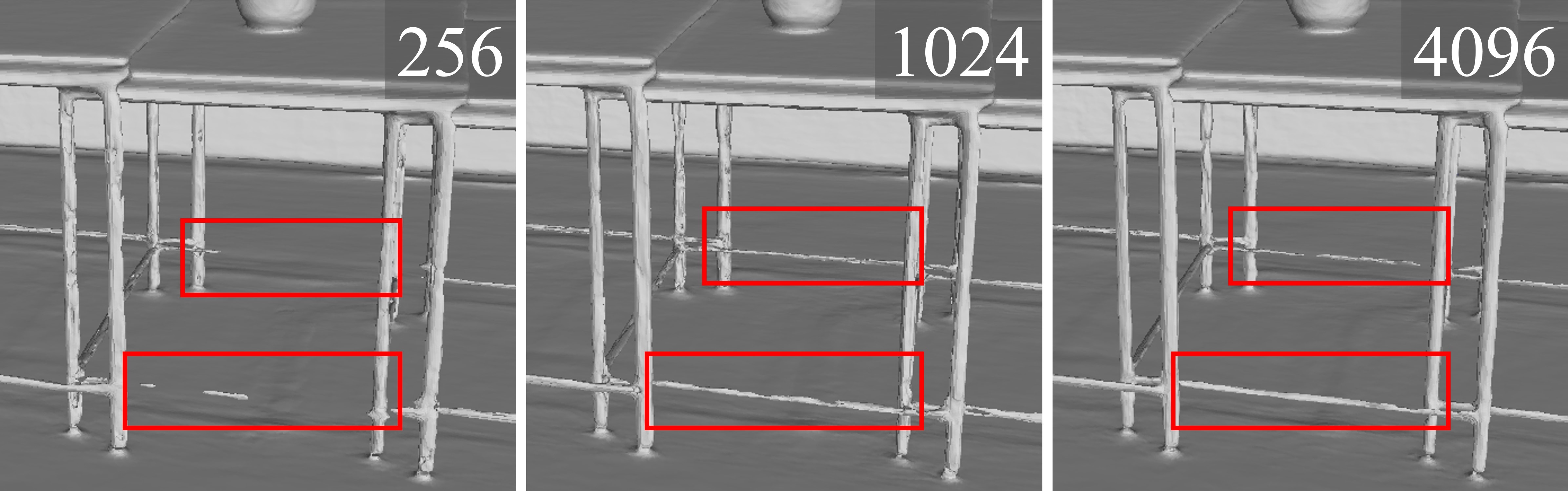}
\end{center}
\vspace{-0.5cm}
   \caption{Reconstruction quality with varying batch size.}
\label{fig:batch_ablation}
\end{figure}

Optimization with a lower batch size leads to more noise and might miss areas without depth supervision due to a lower number of multi-view constraints within the batch.
A batch size that is too large will slow down the optimization and consume more GPU memory, while not offering improvements in reconstruction quality (see Fig.~\ref{fig:batch_ablation}).

\subsection{Truncation Size}

The reconstruction quality is dependent on the width of the truncation region, as shown in Tab.~\ref{tab:truncation}.
The truncation region needs to account for the noise in the input (i.e., needs to be greater than the noise of the depth camera).
In our experiments a truncation radius of $tr=5$~cm gives the best results (evaluated based on the mean across multiple scenes).

\section{Comparison to RGB-based methods}

\begin{figure}[h!]
\begin{center}
\includegraphics[width=\linewidth]{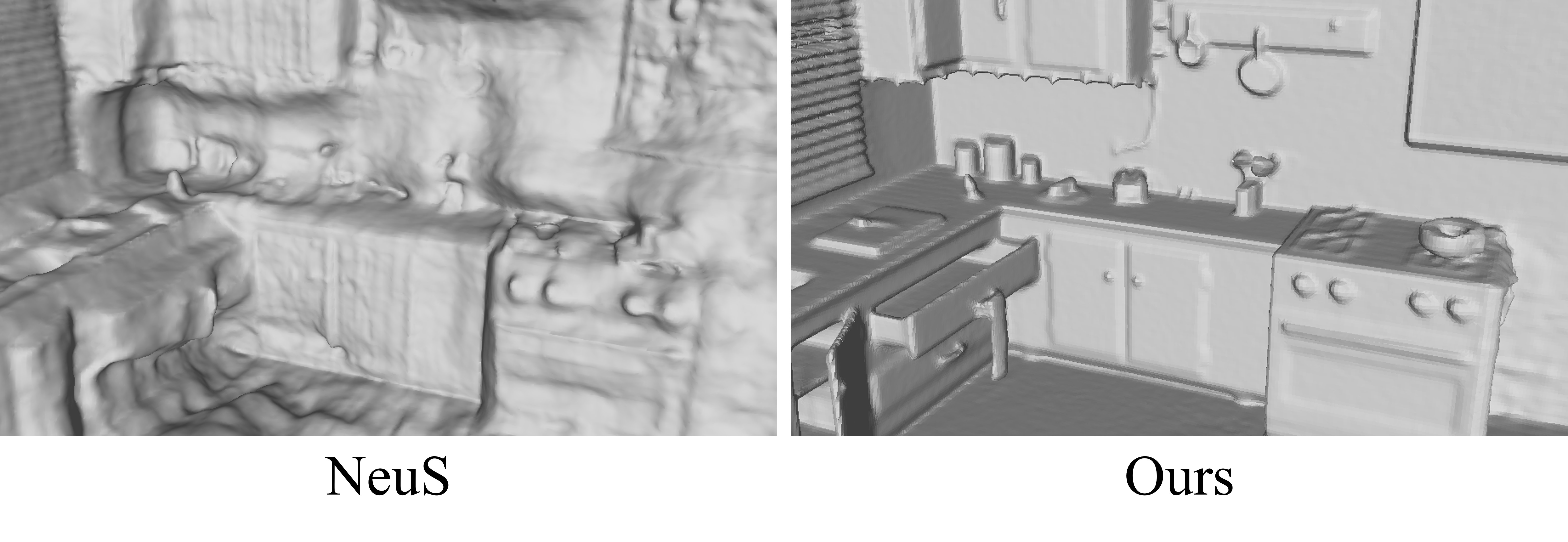}
\end{center}
\vspace{-0.5cm}
   \caption{Comparison between NeuS and our method on the `morning apartment' scene.}
\label{fig:neus_comp}
\end{figure}

NeuS~\cite{wang2021neus} and VolSDF~\cite{yariv2021volume} are concurrent works that propose learning a signed distance field of an object from a set of RGB images.
In contrast to these methods, our focus lies on reconstructing indoor scenes which often have large textureless regions (e.g., a white wall).
Methods which use only color input will not have enough multi-view constraints to properly reconstruct these regions.
In Fig.~\ref{fig:neus_comp}, we show a case where methods that rely only on color input struggle to reconstruct high-quality geometry.

\section{Color Reproduction of Classic and NeRF-style Methods}
\begin{figure}[h!]
\begin{center}
\includegraphics[width=\linewidth]{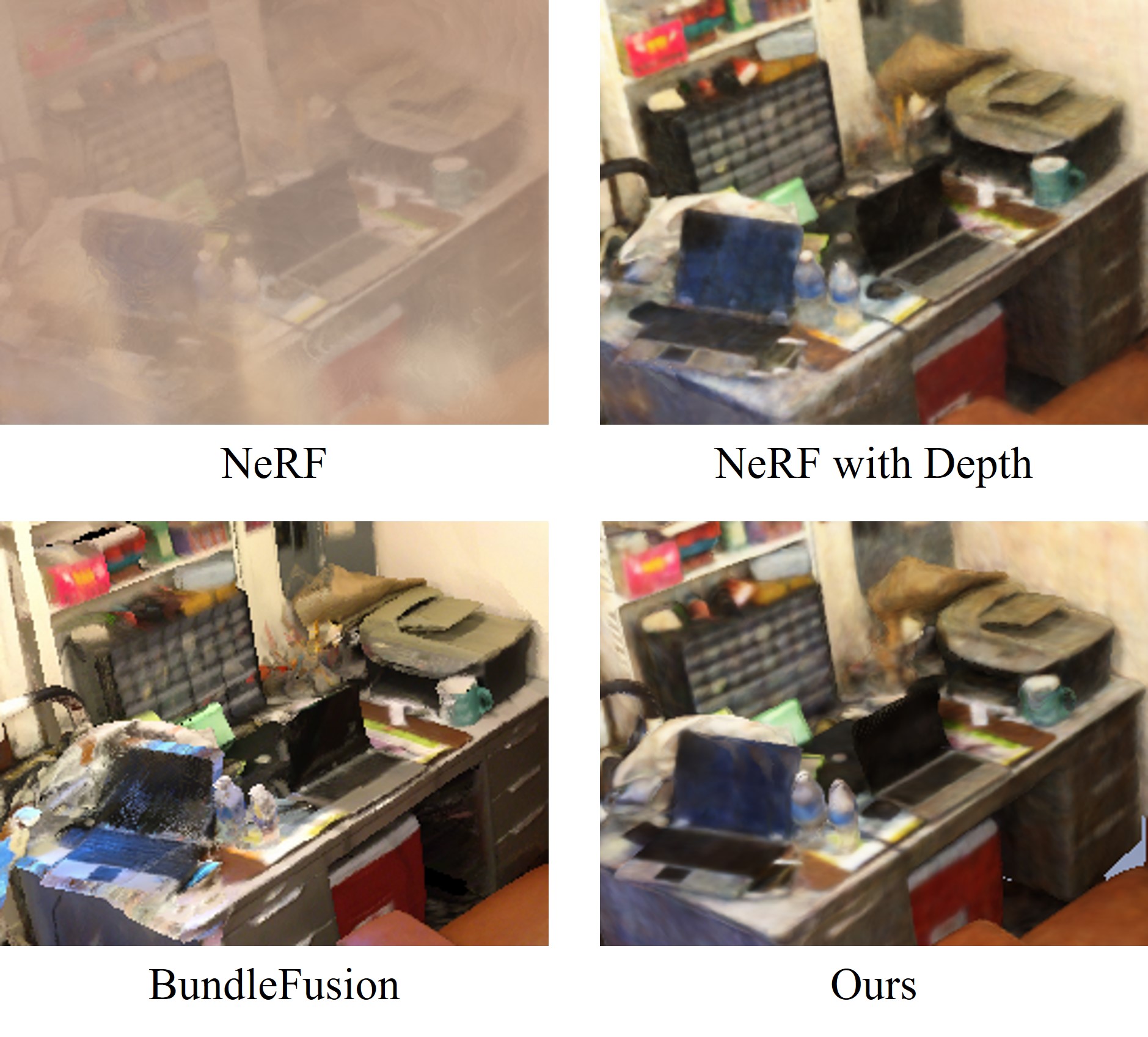}
\end{center}
\vspace{-0.5cm}
   \caption{We compare the color synthesis of BundleFusion and NeRF-style methods. NeRF without any depth constraints shows severe fogging when rendering an image from a novel view. This gets resolved after adding depth constraints to the optimization. BundleFusion produces the sharpest results, but suffers from incorrect view-dependent effects and misalignment artifacts. Our method produces results similar to NeRF with a depth constraint. A combination of classic and NeRF-style methods may yield both high-quality geometry and high-quality view synthesis and we encourage further research in this direction.}
\label{fig:classic_vs_nerf}
\end{figure}

\begin{figure*}[ht!]
\begin{center}
\includegraphics[width=\linewidth]{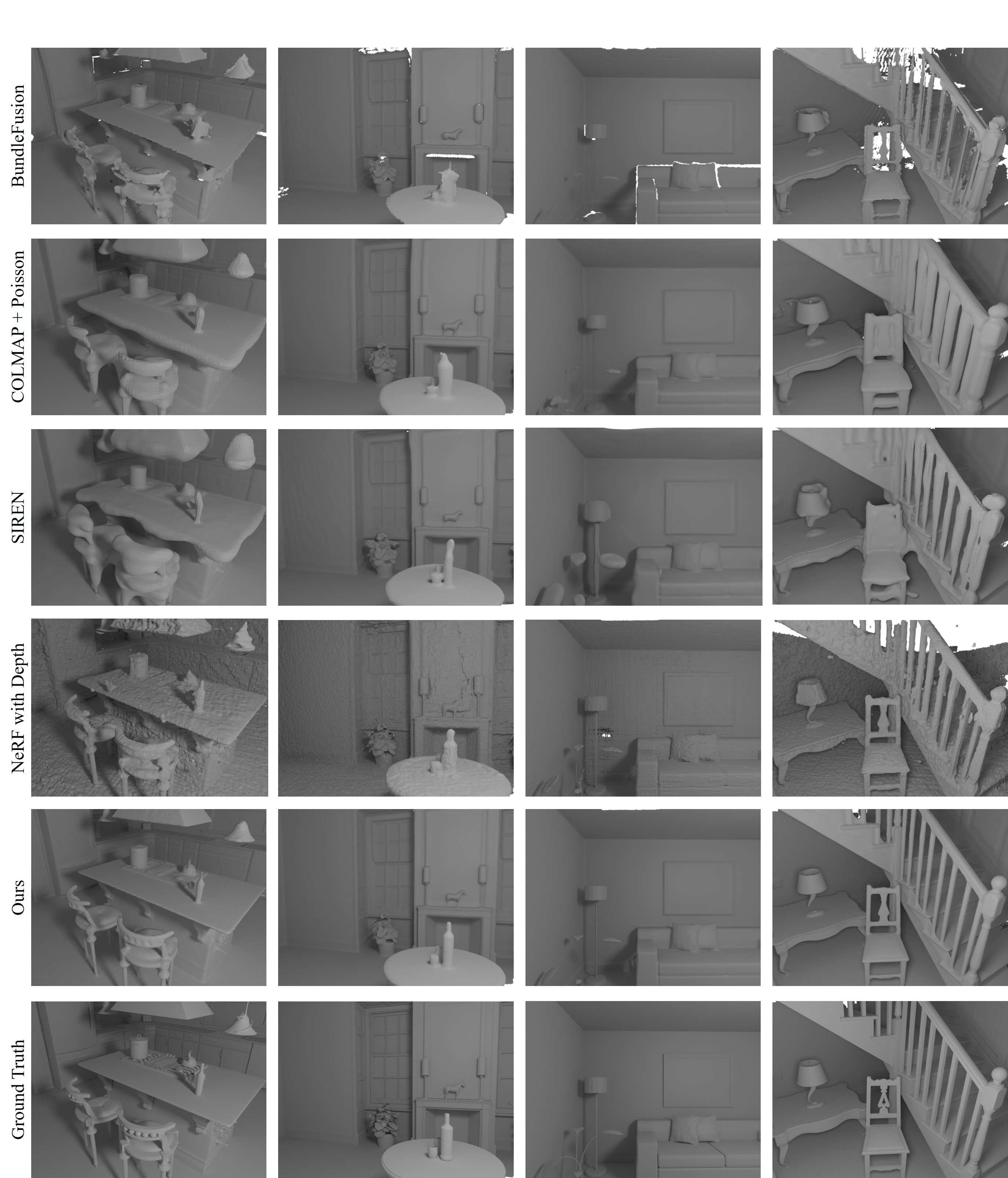}
\end{center}
   \caption{We show a qualitative comparison of synthetic scene reconstructions obtained using our method and several baseline methods. The BundleFusion reconstruction is incomplete in some regions, screened Poisson and SIREN attempt to fit noise in the depth data, while the NeRF reconstruction suffers from noise in the density field. Our method manages to fill in gaps in geometry, while maintaining the smoothness of classic fusion approaches.}
\vspace{-0.2cm}
\label{fig:synth_comparison_1}
\end{figure*}

\begin{figure*}[ht!]
\begin{center}
\includegraphics[width=\linewidth]{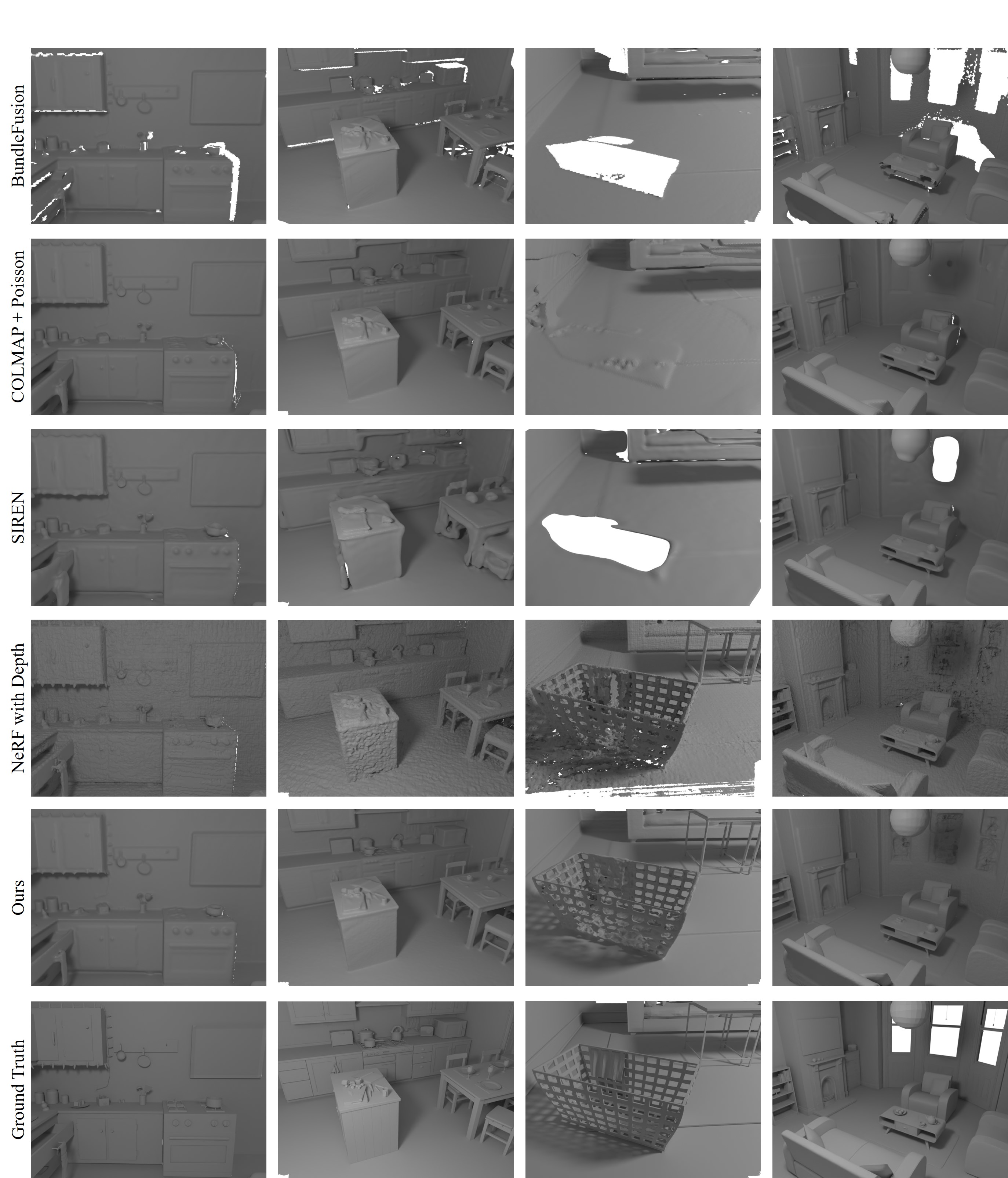}
\end{center}
   \caption{We show a qualitative comparison of synthetic scene reconstructions obtained using our method and several baseline methods. The BundleFusion reconstruction is incomplete in some regions, screened Poisson and SIREN attempt to fit noise in the depth data, while the NeRF reconstruction suffers from noise in the density field. Our method manages to fill in gaps in geometry, while maintaining the smoothness of classic fusion approaches.}
\vspace{-0.2cm}
\label{fig:synth_comparison_2}
\end{figure*}

While our focus lies on geometry reconstruction and not accurate view synthesis, we conducted a brief analysis of the advantages and drawbacks of classic reconstruction methods~\cite{dai2017bundlefusion,color_map} and MLP-based radiance fields~\cite{mildenhall2020nerf} when synthesizing unseen views.
Classic reconstruction methods usually do not try to decouple intrinsic material parameters~\cite{dai2017bundlefusion,color_map} and instead optimize a texture that represents the average observation of all the input views.
The resulting texture is usually high-resolution (bounded by the resolution of the input images), but does not allow for correct synthesis of view-dependent effects.
Furthermore, inaccuracies in camera calibration may lead to visible seams in the optimized texture.
Methods like NeRF that focus purely on high-quality novel view synthesis do not explicitly reconstruct geometry and may thus produce images riddled with artifacts for views that are too far from the input views.
We believe that it is possible to combine both of these approaches to improve novel view synthesis on views far away from the ones used during the optimization and would like to encourage research in this direction.
Fig.~\ref{fig:classic_vs_nerf} shows an example view synthesis result on the ScanNet dataset, for an out-of-trajectory camera position and orientation.

\section{Runtime and Memory Requirements}
\label{sec:timings}

\paragraph{Our method.}
The runtime and memory requirements of our method are dependent on the scene size.
For smaller scenes where it is enough to have $S'_c=256$ samples, our method completes $2\times10^5$ iterations in $9$ hours on an NVIDIA RTX 3090 and requires $8.5$~GB of GPU memory.
When $S'_c$ is set to 512, the runtime increases to $13$ hours and the memory requirement to $10.5$~GB.
The memory consumption can be reduced by using smaller batches.

\paragraph{BundleFusion.}
We run BundleFusion at a voxel resolution of $1$~cm for all scenes.
On an NVIDIA GTX TITAN Black, depending on the size of the scene and number of frames in the camera trajectory, it takes $10$ to $40$ minutes to integrate the depth frames into a truncated signed distance field and extract a mesh using Marching Cubes.
The memory usage is around $5.8$~GB.

\paragraph{RoutedFusion.}
To train and test RoutedFusion, we used an NVIDIA RTX 3090.
The routing network was trained for 24 hours on images with a resolution of $320\times240$ pixels.
As per suggestion of the authors, we train the fusion network for $20$ epochs which takes about $1.5$ hours.
We reconstruct all scenes at a voxel resolution of $1$~cm for a fair comparison to other methods.
The runtime ranges from $40$ minutes to $6$ hours depending on scene size and number of frames.
The memory usage also heavily depends on scene size and ranges from $5.5$~GB to $23$~GB.

\paragraph{COLMAP + Poisson.}
In the COLMAP + Poisson baseline, the bottleneck is the global bundle adjustment process performed by COLMAP.
The total runtime depends on the number of frames in the trajectory.
Using all 8 cores of an Intel i7-7700K CPU, it took us about $4$ hours to align all 1167 cameras in the `breakfast room'.
The couple of minutes needed to backproject all depth maps at full resolution and run the screened Poisson surface reconstruction are negligible in comparison.

\paragraph{Convolutional Occupancy Networks.}
We reconstruct each scene using the pre-trained model provided by the authors.
This takes about $2$ minutes per scene and requires about $10$~GB of memory.

\paragraph{SIREN.}
We train SIREN for $10^4$ epochs on each scene.
SIREN is trained over the complete point cloud in each epoch, so the runtime depends on the number of points in the point cloud.
In our experiments on an NVIDIA RTX 3090, this ranged from $6$ to $12$ hours with $12$~GB of memory being in use.

\paragraph{NeRF + Depth.}
We optimize NeRF using 64 samples for the coarse network and 128 samples for the fine network.
On an NVIDIA RTX 3090 it takes $6$ hours for $2\times10^5$ iterations to run.
The memory usage is $4.7$~GB.

\clearpage
\onecolumn
\begin{longtable}{llcccccc}
        \toprule
        \textbf{Scene}             & \textbf{Method}  & \textbf{C-$\ell_1$} $\downarrow$  & \textbf{IoU} $\uparrow$ & \textbf{NC} $\uparrow$  & \textbf{F-score} $\uparrow$ & \textbf{Pos. error} $\downarrow$ & \textbf{Rot. error} $\downarrow$ \\
        \midrule   
        \textbf{Breakfast room}    & BundleFusion     & 0.033          & 0.698          & \textbf{0.944} & 0.890          & 0.037          & 0.697              \\
        \multirow{7}{*}[-0.1cm]{\includegraphics[width=1.5in]{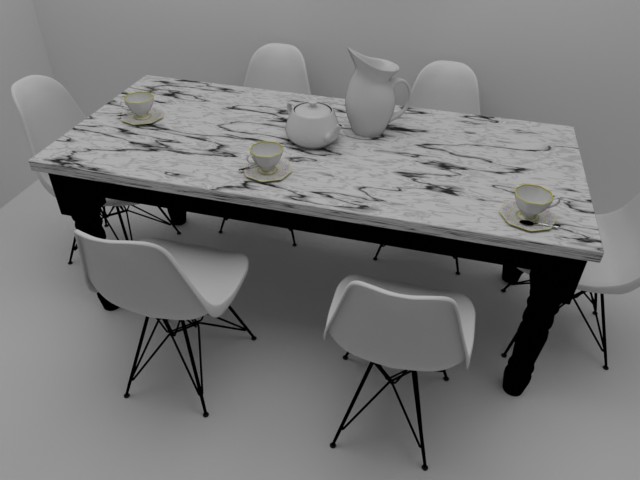}}
                                   & RoutedFusion     & 0.033          & 0.714          & 0.918          & 0.901          &  -              &  -                  \\
                                   & COLMAP + Poisson & 0.033          & 0.668          & 0.935          & 0.893          & 0.009          & 0.210              \\
                                   & Conv. Occ. Nets  & 0.047          & 0.474          & 0.879          & 0.780          &   -             &  -                  \\
                                   & SIREN            & 0.060          & 0.566          & 0.922          & 0.822          & -               &  -                  \\
                                   & NeRF + Depth     & 0.041          & 0.619          & 0.811          & 0.854          &  -              &  -                  \\
        \cmidrule{2-8}       
                                   & Ours (w/o pose)  & 0.031          & 0.720          & 0.930          & 0.914          &   -             &   -                 \\
                                   & Ours             & \textbf{0.030} & \textbf{0.793} & 0.934          & \textbf{0.920} & \textbf{0.007} & \textbf{0.135}     \\
        \midrule
        \textbf{Green room}        & BundleFusion     & 0.024          & 0.694          & 0.923          & 0.926          & 0.027          & 0.546              \\
        \multirow{7}{*}[-0.1cm]{\includegraphics[width=1.5in]{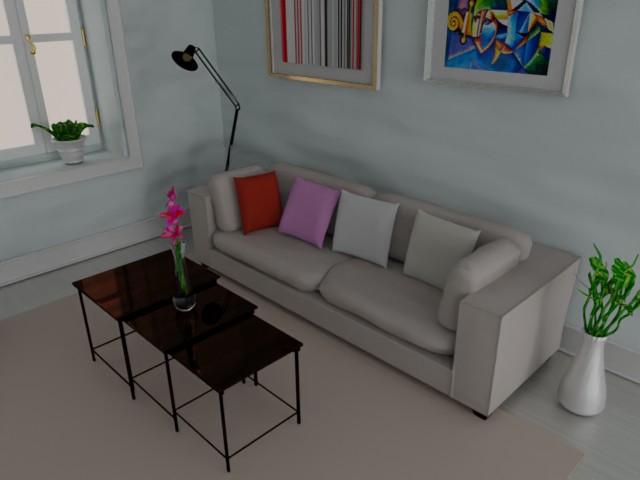}}
                                   & RoutedFusion     & 0.018          & 0.755          & 0.904          & 0.969          &   -             &   -                 \\
                                   & COLMAP + Poisson & 0.018          & 0.849          & 0.925          & 0.967          & 0.014          & 0.227              \\
                                   & Conv. Occ. Nets  & 0.053          & 0.554          & 0.855          & 0.737          &  -              &    -                \\
                                   & SIREN            & 0.023          & 0.746          & 0.913          & 0.940          &  -              &   -                 \\
                                   & NeRF + Depth     & 0.030          & 0.668          & 0.748          & 0.871          &   -             &    -                \\
        \cmidrule{2-8}         
                                   & Ours (w/o pose)  & 0.014          & 0.766          & 0.931          & 0.982          &  -              &    -                \\
                                   & Ours             & \textbf{0.013} & \textbf{0.921} & \textbf{0.932} & \textbf{0.990} & \textbf{0.012} & \textbf{0.104}     \\
        \midrule         
        \textbf{Grey-white room}   & BundleFusion     & 0.038          & 0.567          & 0.860          & 0.751          & 0.056          & 1.891              \\
        \multirow{7}{*}[-0.1cm]{\includegraphics[width=1.5in]{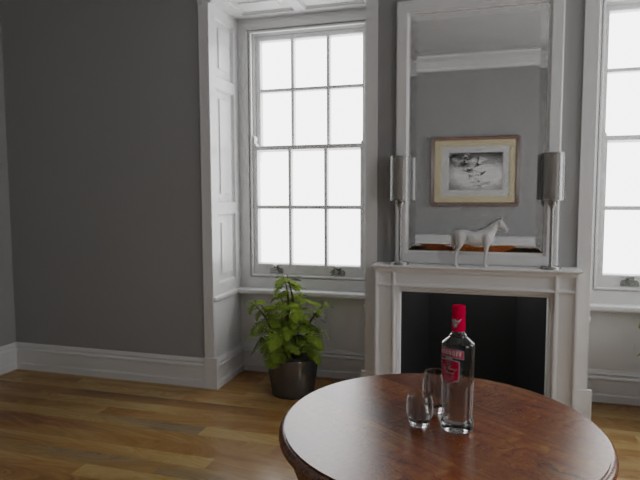}}
                                   & RoutedFusion     & 0.033          & 0.606          & 0.850          & 0.790          &  -              &  -                  \\
                                   & COLMAP + Poisson & 0.029          & 0.727          & 0.899          & 0.899          & 0.029          & 0.296              \\
                                   & Conv. Occ. Nets  & 0.048          & 0.480          & 0.841          & 0.601          &  -              &    -                \\
                                   & SIREN            & 0.033          & 0.635          & 0.868          & 0.812          &   -             &   -                 \\
                                   & NeRF + Depth     & 0.040          & 0.563          & 0.764          & 0.697          &    -            &   -                 \\
        \cmidrule{2-8}         
                                   & Ours (w/o pose)  & 0.032          & 0.640          & 0.864          & 0.806          &  -              &    -                \\
                                   & Ours             & \textbf{0.015} & \textbf{0.886} & \textbf{0.924} & \textbf{0.987} & \textbf{0.014} & \textbf{0.146}     \\
        \midrule         
        \textbf{ICL living room}   & BundleFusion     & 0.018          & 0.743          & 0.956          & 0.958          & 0.022          & 0.382              \\
        \multirow{7}{*}[-0.1cm]{\includegraphics[width=1.5in]{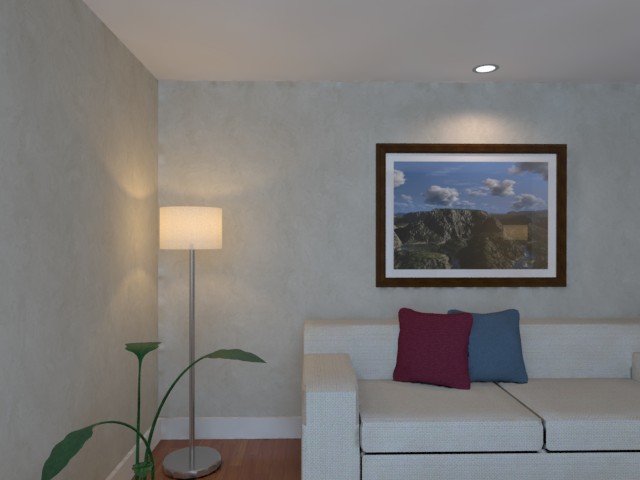}}
                                   & RoutedFusion     & 0.019          & 0.698          & 0.939          & 0.976          &  -              &  -                  \\
                                   & COLMAP + Poisson & 0.023          & 0.727          & 0.947          & 0.966          & 0.029          & 0.836              \\
                                   & Conv. Occ. Nets  & 0.112          & 0.352          & 0.841          & 0.507          & -               &  -                  \\
                                   & SIREN            & 0.020          & 0.768          & 0.950          & 0.967          &  -              &  -                  \\
                                   & NeRF + Depth     & 0.021          & 0.689          & 0.900          & 0.956          &   -             &   -                 \\
        \cmidrule{2-8}         
                                   & Ours (w/o pose)  & 0.014          & 0.790          & 0.964          & 0.992          &  -              &  -                  \\
                                   & Ours             & \textbf{0.011} & \textbf{0.905} & \textbf{0.969} & \textbf{0.994} & \textbf{0.007} & \textbf{0.109}     \\
        \midrule         
        \textbf{Kitchen 1}         & BundleFusion     & \textbf{0.234} & 0.368          & 0.860          & 0.620          & 0.038          & 0.327              \\
        \multirow{7}{*}[-0.1cm]{\includegraphics[width=1.5in]{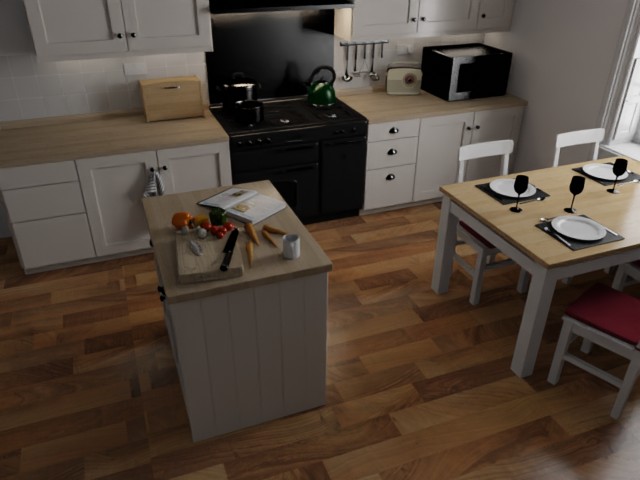}}
                                   & RoutedFusion     & 0.265          & 0.401          & 0.805          & 0.680          &  -              &    -                \\
                                   & COLMAP + Poisson & 0.252          & \textbf{0.459} & \textbf{0.888} & \textbf{0.748} & 0.103          & 0.941              \\
                                   & Conv. Occ. Nets  & 0.262          & 0.352          & 0.839          & 0.483          &  -              &   -                 \\
                                   & SIREN            & 0.265          & 0.357          & 0.850          & 0.575          &   -             &    -                \\
                                   & NeRF + Depth     & 0.271          & 0.336          & 0.710          & 0.600          &   -             &    -                \\
        \cmidrule{2-8}         
                                   & Ours (w/o pose)  & 0.255          & 0.420          & 0.887          & 0.700          &  -              &   -                 \\
                                   & Ours             & 0.252          & 0.447          & 0.886          & 0.718          & \textbf{0.030} & \textbf{0.114}     \\
        \bottomrule
    \caption{We compare the quality of our reconstruction on several synthetic scenes for which ground truth data is available. The Chamfer $\ell_1$ distance, normal consistency and the F\mbox{-}score~\cite{10.1145/3072959.3073599} are computed between point clouds sampled with a density of 1 point per cm\textsuperscript{2}. We use a threshold of 5~cm for the F-score. We further voxelize each mesh to compute the intersection-over-union (IoU) between the predictions and ground truth.}
\vspace{-0.3cm}
\label{tab:quantitative_per_scene_1}
\end{longtable}      
\clearpage

\begin{longtable}{llcccccc}
        \toprule
        \textbf{Scene}             & \textbf{Method}  & \textbf{C-$\ell_1$} $\downarrow$  & \textbf{IoU} $\uparrow$ & \textbf{NC} $\uparrow$  & \textbf{F-score} $\uparrow$ & \textbf{Pos. error} $\downarrow$ & \textbf{Rot. error} $\downarrow$ \\
        \midrule           
        
        \textbf{Kitchen 2}         & BundleFusion     & 0.089          & 0.441          & 0.856          & 0.687          & 0.050          & 0.566              \\
        \multirow{7}{*}[-0.1cm]{\includegraphics[width=1.5in]{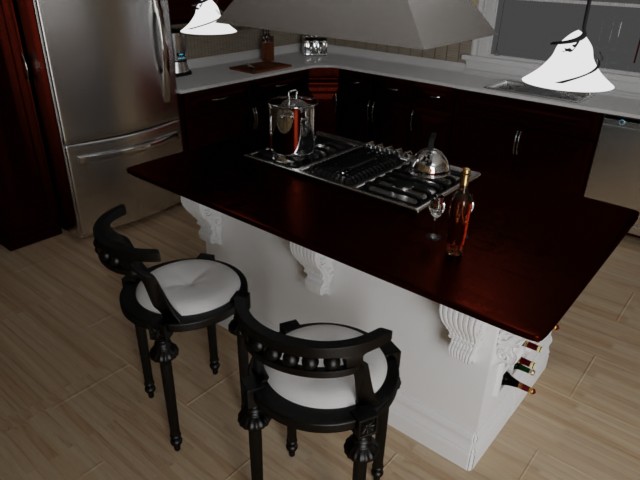}}
                                   & RoutedFusion     & 0.059          & 0.572          & 0.842          & 0.787          &   -             &   -                 \\
                                   & COLMAP + Poisson & 0.037          & \textbf{0.675} & \textbf{0.919} & 0.818          & \textbf{0.043} & 1.154              \\
                                   & Conv. Occ. Nets  & 0.052          & 0.484          & 0.861          & 0.653          &  -              &    -                \\
                                   & SIREN            & 0.055          & 0.453          & 0.898          & 0.735          &   -             &       -             \\
                                   & NeRF + Depth     & 0.051          & 0.435          & 0.708          & 0.630          &  -              &    -                \\
        \cmidrule{2-8}         
                                   & Ours (w/o pose)  & 0.034          & 0.488          & 0.908          & 0.796          &  -              &    -                \\
                                   & Ours             & \textbf{0.032} & 0.637          & 0.903          & \textbf{0.890} & 0.083          & \textbf{0.450}     \\
        \midrule
        \textbf{Morning apartment} & BundleFusion     & 0.012          & 0.767          & 0.885          & 0.968          & 0.008          & 0.165              \\
        \multirow{7}{*}[-0.1cm]{\includegraphics[width=1.5in]{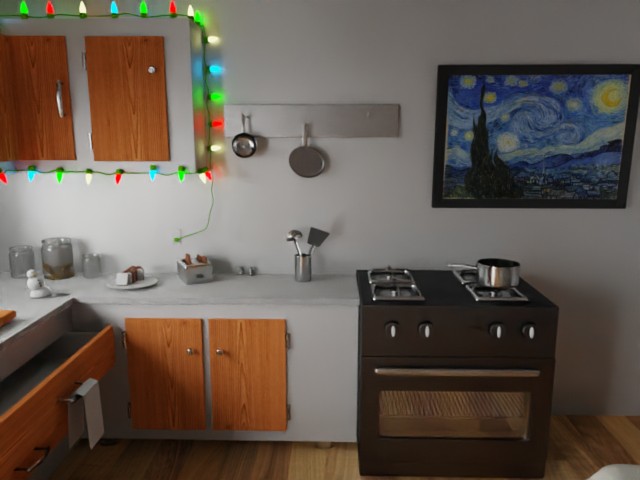}}
                                   & RoutedFusion     & 0.013          & \textbf{0.815} & 0.870          & 0.976          &   -             &     -               \\
                                   & COLMAP + Poisson & 0.017          & 0.668          & 0.877          & 0.959          & 0.017          & 0.380              \\
                                   & Conv. Occ. Nets  & 0.045          & 0.450          & 0.802          & 0.784          &  -              &     -               \\
                                   & SIREN            & 0.013          & 0.727          & 0.873          & 0.966          &   -             &     -               \\
                                   & NeRF + Depth     & 0.022          & 0.587          & 0.838          & 0.975          &  -              &    -                \\
        \cmidrule{2-8}         
                                   & Ours (w/o pose)  & \textbf{0.011} & 0.787          & 0.887          & \textbf{0.983} &        -        &      -              \\
                                   & Ours             & \textbf{0.011} & 0.716          & \textbf{0.888} & 0.982          & \textbf{0.005} & \textbf{0.093}     \\
        \midrule         
        \textbf{Staircase}         & BundleFusion     & 0.091          & 0.373          & 0.860          & 0.623          & 0.039          & 0.643              \\
        \multirow{7}{*}[-0.1cm]{\includegraphics[width=1.5in]{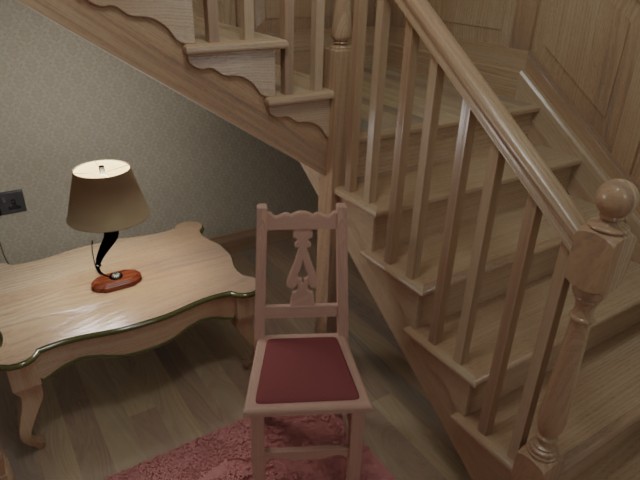}}
                                   & RoutedFusion     & 0.069          & 0.340          & 0.864          & 0.622          &  -              &    -                \\
                                   & COLMAP + Poisson & 0.074          & 0.322          & 0.895          & 0.628          & 0.043          & 0.305              \\
                                   & Conv. Occ. Nets  & 0.069          & 0.315          & 0.838          & 0.508          &  -              &    -                \\
                                   & SIREN            & 0.067          & 0.432          & 0.885          & 0.676          &  -              &    -                \\
                                   & NeRF + Depth     & 0.087          & 0.396          & 0.644          & 0.624          &  -              &   -                 \\
        \cmidrule{2-8}         
                                   & Ours (w/o pose)  & 0.057          & 0.457          & 0.899          & 0.704          &   -             &    -                \\
                                   & Ours             & \textbf{0.045} & \textbf{0.565} & \textbf{0.920} & \textbf{0.853} & \textbf{0.016} & \textbf{0.123}     \\
        \midrule
        \textbf{Thin geometry}     & BundleFusion     & 0.019          & 0.764          & 0.909          & 0.922          & \textbf{0.009} & 0.126              \\
        \multirow{7}{*}[-0.1cm]{\includegraphics[width=1.5in]{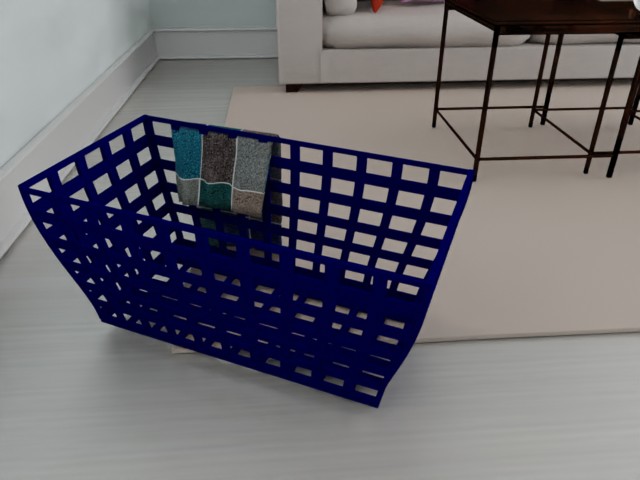}}
                                   & RoutedFusion     & 0.023          & 0.708          & 0.829          & 0.881          &  -              &    -                \\
                                   & COLMAP + Poisson & 0.047          & 0.440          & 0.820          & 0.721          & 0.079          & 2.400              \\
                                   & Conv. Occ. Nets  & 0.022          & 0.723          & 0.882          & 0.910          &   -             &    -                \\
                                   & SIREN            & 0.021          & 0.733          & 0.887          & 0.913          &   -             &   -                 \\
                                   & NeRF + Depth     & 0.014          & 0.825          & 0.847          & 0.989          &   -             &   -                 \\
        \cmidrule{2-8}
                                   & Ours (w/o pose)  & \textbf{0.009} & 0.857          & \textbf{0.911} & \textbf{0.995} &           -     &      -              \\
                                   & Ours             & \textbf{0.009} & \textbf{0.865} & 0.910          & \textbf{0.995} & 0.010          & \textbf{0.037}     \\
        \midrule
        \textbf{White room}        & BundleFusion     & 0.062          & 0.528          & 0.869          & 0.701          & 0.045          & 0.375              \\
        \multirow{7}{*}[-0.1cm]{\includegraphics[width=1.5in]{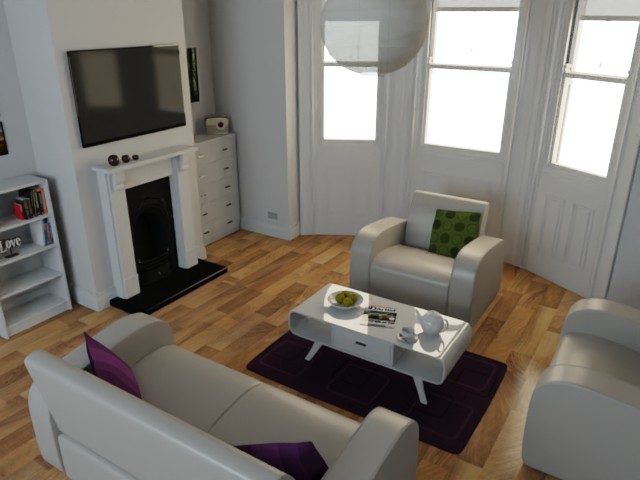}}
                                   & RoutedFusion     & 0.038          & 0.545          & 0.817          & 0.799          &        -        &           -         \\
                                   & COLMAP + Poisson & 0.036          & 0.652          & 0.904          & 0.796          & \textbf{0.018} & 0.167              \\
                                   & Conv. Occ. Nets  & 0.061          & 0.424          & 0.853          & 0.470          &        -        &        -            \\
                                   & SIREN            & 0.046          & 0.617          & 0.888          & 0.752          &        -        &          -          \\
                                   & NeRF + Depth     & 0.073          & 0.385          & 0.716          & 0.619          &       -         &       -             \\
        \cmidrule{2-8}
                                   & Ours (w/o pose)  & 0.034          & 0.631          & 0.902          & 0.813          &       -         &       -             \\
                                   & Ours             & \textbf{0.028} & \textbf{0.738} & \textbf{0.911} & \textbf{0.915} & 0.028          & \textbf{0.133}     \\
        \bottomrule
    \caption{We compare the quality of our reconstruction on several synthetic scenes for which ground truth data is available. The Chamfer $\ell_1$ distance, normal consistency and the F\mbox{-}score~\cite{10.1145/3072959.3073599} are computed between point clouds sampled with a density of 1 point per cm\textsuperscript{2}. We use a threshold of 5~cm for the F-score. We further voxelize each mesh to compute the intersection-over-union (IoU) between the predictions and ground truth.}
\vspace{-0.3cm}
\label{tab:quantitative_per_scene_2}
\end{longtable}

\clearpage
\twocolumn

\newpage

\end{appendix}

\end{document}